\documentclass[11pt]{article}

\usepackage[final]{acl}

\usepackage{times}
\usepackage{latexsym}
\usepackage{iftex}
\ifPDFTeX
  \usepackage[LAE,T1]{fontenc}
  \usepackage[utf8]{inputenc}
  \usepackage[arabic,english]{babel} 
  \usepackage{CJKutf8}
  \newcommand{\zh}[1]{\begin{CJK}{UTF8}{gbsn}#1\end{CJK}}
  \DeclareUnicodeCharacter{2248}{\ensuremath{\approx}}
  \newcommand{\arabictext}[1]{\AR{#1}} 
\else
  \usepackage[T1]{fontenc}
  \usepackage[english,bidi=default]{babel} 
  \babelfont{rm}{TeXGyreTermesX} 
  \babelprovide[import]{arabic}
  \babelfont[*arabic]{rm}{Noto Sans Arabic}
  \usepackage{xeCJK}
  \newcommand{\zh}[1]{#1}
  \newcommand{\arabictext}[1]{\foreignlanguage{arabic}{#1}}
\fi
\usepackage{xcolor}
\usepackage{colortbl}

\definecolor{gold}{RGB}{255, 215, 0}
\definecolor{silver}{RGB}{192, 192, 192}
\definecolor{bronze}{RGB}{205, 127, 50}
\definecolor{fourth}{RGB}{176, 224, 230}  

\newcommand{\gld}[1]{\cellcolor{gold!40}\textbf{#1}}
\newcommand{\slv}[1]{\cellcolor{silver!50}#1}
\newcommand{\brz}[1]{\cellcolor{bronze!30}#1}
\newcommand{\fth}[1]{\cellcolor{fourth!60}#1}

\usepackage{microtype}

\usepackage{amssymb}
\usepackage{graphicx}
\usepackage{subcaption}
\usepackage{makecell}
\usepackage{booktabs}
\usepackage{multirow}
\usepackage{tcolorbox}
\usepackage{listings}

\usepackage{siunitx}
\definecolor{yamlkey}{rgb}{0.25,0.5,0.75}
\definecolor{yamlstring}{rgb}{0,0.5,0}
\definecolor{yamlcomment}{rgb}{0.5,0.5,0.5}

\lstdefinelanguage{yaml}{
  keywords={true,false,null},
  keywordstyle=\color{yamlkey}\bfseries,
  comment=[l]{\#},
  commentstyle=\color{yamlcomment}\ttfamily,
  stringstyle=\color{yamlstring},
  morestring=[b]",
  morestring=[b]'
}

\lstdefinestyle{yamlstyle}{
  language=yaml,
  basicstyle=\ttfamily\footnotesize,
  frame=single,
  breaklines=true,
  showstringspaces=false,
  columns=flexible,
  xleftmargin=2pt,
  xrightmargin=2pt,
  literate={→}{{$\rightarrow$}}1 {é}{{\'e}}1
}

\title{Multicultural Spyfall: Assessing LLMs through a Dynamic Multilingual Social Deduction Game}

\author{
Haryo Akbarianto Wibowo$^{1}$, Alaa Elsetohy$^{1}$, Qinrong Cui$^{1}$, Alham Fikri Aji$^1$\\
$^{1}$MBZUAI \\
\texttt{\small \{haryo.wibowo,alaa.elsetohy,qinrong.cui,alham.fikri\}@mbzuai.ac.ae}
}

\newcommand{\red}[0]{}

\begin{document}
\maketitle

\begin{abstract}
    The rapid advancement of Large Language Models (LLMs) has necessitated more robust evaluation methods that go beyond static benchmarks, which are increasingly prone to data saturation and leakage. In this paper, we propose a dynamic benchmarking framework for evaluating multilingual and multicultural capabilities through the social deduction game Spyfall. In our setup, models must engage in strategic dialogue to either identify a secret agent or avoid detection, utilizing culturally relevant locations or local foods. \red{Our results show that our game-based rankings are consistent with Chatbot Arena rankings on the overlapping models}. However, we find a significant performance gap in non-English contexts: models are generally less proficient when handling locally specific entities and often struggle with rule-following or strategic integrity in non-English languages. We demonstrate that this game-based approach provides a scalable, leakage-resistant, and culturally nuanced alternative to traditional NLP benchmarks. The dataset (game history) and source code can be accessed here\footnote{\url{https://huggingface.co/datasets/haryoaw/cultural-spyfall} and \url{https://github.com/haryoa/cultural-spyfall}}.
\end{abstract}

\section{Introduction}
The rapid advancement of Large Language Models (LLMs) and their expanding multilingual capabilities have made robust evaluation increasingly critical. While numerous multilingual benchmarks have been developed~\cite{wu2025bitterlessonlearned2000,hu2020xtreme}, many are static, making them susceptible to data saturation and potential ``leakage'' into training sets over time. To address these limitations, researchers have explored dynamic benchmarking, utilizing text-based games~\cite{hugamearena, song2025survivalevaluatingllmssocial,ma2025hungergamedebateemergence,kim2025fine}, human preference evaluations~\cite{chiang2024chatbotarenaopenplatform, kim2025fine}, or debates~\cite{moniri-etal-2025-debate}. However, a significant gap remains in combining these dynamic approaches to specifically evaluate multilingual and multicultural nuances.
  
To bridge this gap, we propose a dynamic benchmarking framework for multilingual and multicultural understanding through strategic gameplay. Specifically, we adapt the social deduction game Spyfall\footnote{\url{https://hobbyworldint.com/portfolio-item/spyfall/}}. In a standard game of Spyfall, all players except one (the ``Spy'') are given a specific location; the Spy's goal is to deduce that location through conversation, while the other players attempt to identify the Spy by asking subtle, context-specific questions.

\begin{figure}
    \centering
    \includegraphics[width=1\linewidth]{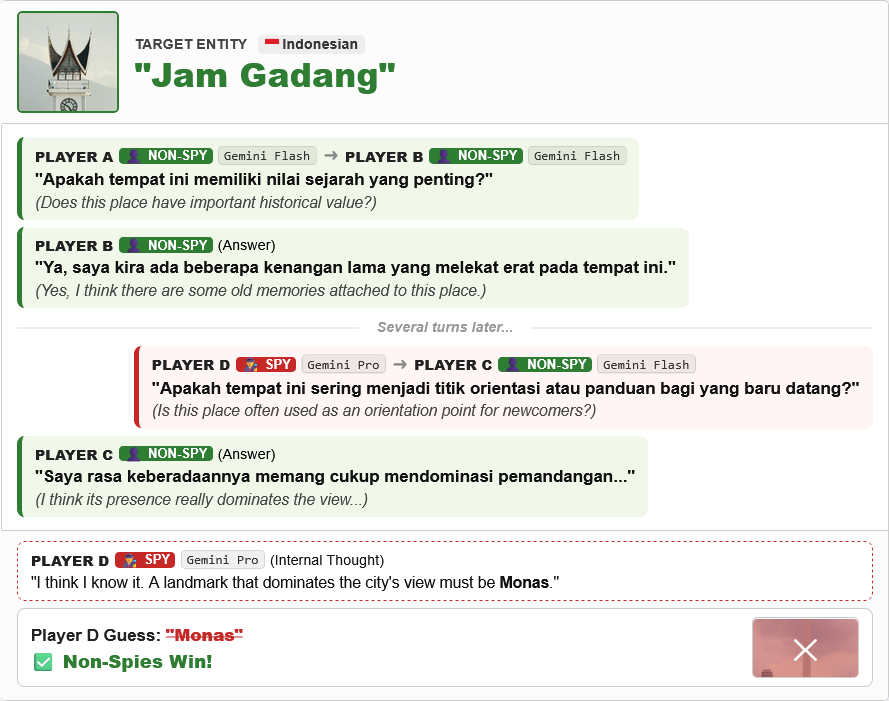}
    \caption{In this example, the Spy fails to identify the target entity, Jam Gadang. Due to insufficient context and careful questioning from the other players, the Spy guesses Monas (a landmark in a different part of Indonesia) and loses the game.}
    \label{fig:hook}
\end{figure}
\begin{figure*}[htbp]
    \centering
    \includegraphics[width=\linewidth]{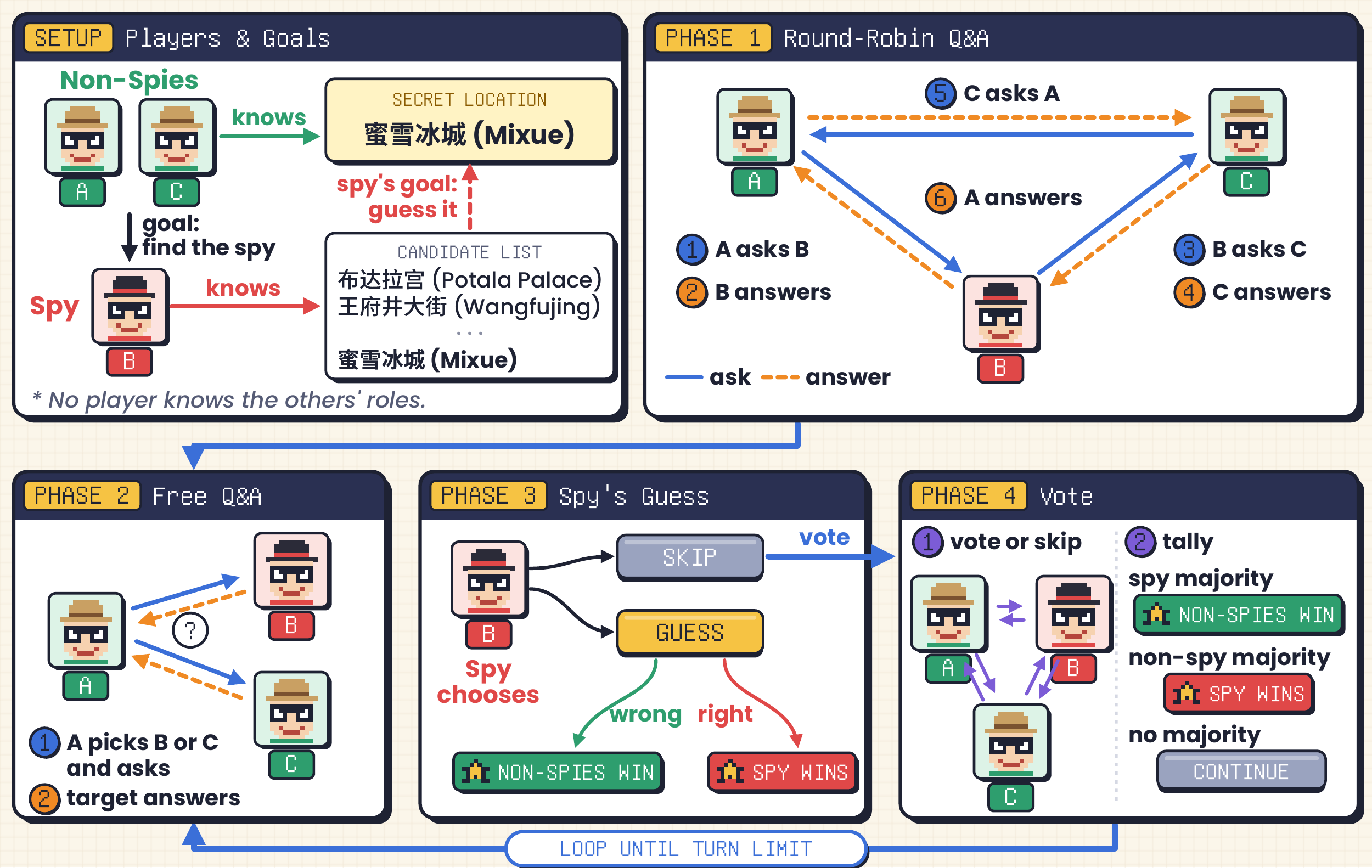}
    \caption{The flow of the game of Multicultural Spyfall}
    \label{fig:game-flow}
\end{figure*}

Our framework utilizes this concept by playing the game in diverse languages and replacing generic locations with culturally relevant settings. This requires models to not only understand the language but also possess a deep cultural understanding, in our case, location and food as a proxy for culture, following~\cite{adilazuarda-etal-2024-towards}.

Our findings indicate that our benchmark rankings follow Chatbot Arena's ordering of the models we evaluate, though our rating spread is wider. Furthermore, our analysis reveals that model proficiency significantly declines in non-English contexts and when handling locally specific cultural entities. We observed that while top-tier models remain competitive, weaker models often fail to follow game logic or inadvertently leak the secret entity to the spy when operating in non-English languages. We also find that guessing food is more challenging than guessing location.
  
By utilizing social deduction games to evaluate multilingual and multicultural capabilities, our benchmark addresses several flaws in current evaluation methodologies. First, it is inherently resistant to saturation; because models compete against one another in a dynamic environment, the `performance ceiling' evolves as models improve. Second, it is robust against test set data leakage; that is, it does not rely on the dataset due to the interactive nature of the evaluation system. Finally, this framework is easy to scale and is extensible; localized versions for new languages or cultures can be implemented simply by updating the underlying entity pools and cultural settings, thus eliminating the need for the labor-intensive manual annotation typically required for new benchmarks.

In summary, our contributions are as follows:
\begin{itemize}
    \item We introduce a dynamic benchmarking framework by extending the social deduction game Spyfall to evaluate the multilingual and multicultural reasoning capabilities of LLMs.
    \item We show that, due to the nature of the proposed framework, it not only assesses the overall capability of each model in this game (validated by comparing its ranking with that of the LMSYS Chatbot Arena) but also provides a comprehensive analysis of each model across different entangled axes: language proficiency, cultural knowledge, and strategic reasoning. For instance, we identify performance degradation in non-English settings and in locally specific cultural contexts. We also find the framework useful for revealing how each model behaves in this game.
\end{itemize}


\section{Multicultural Spyfall}

We adapt the board game Spyfall into a turn-based format to accommodate LLM benchmarking. Moreover, we expand the game to support multiple languages and extend the set of entities from generic locations to locally specific locations and foods. We call this framework Multicultural Spyfall.

\subsection{Game Mechanics}

The game is played by 3 to 8 players and features two primary roles: \textbf{one spy} and the remaining players as \textbf{non-spies}. Each player is given a list of 30 possible entities. At the start of the game, non-spy players are informed of one of the 30 entities, while the spy does not receive any entity information. No player knows the roles of the others. The non-spies aim to identify the spy, whereas the spy attempts to deduce the entity without being detected.

The gameplay proceeds through three phases: a question-and-answer phase ($\mathcal{P}_\text{QA}$), in which players interrogate one another with questions relevant to the given entity; a spy-guessing phase ($\mathcal{P}_\text{SG}$), in which the spy attempts to identify the secret entity; and a voting phase ($\mathcal{P}_\text{V}$), where any player can accuse others (including the spy voting against non-spies). The game's design subtly engages all participants. Specifically, the spy must avoid exposing themselves during questioning, while non-spies must respond without revealing obvious hints about the entity and simultaneously deduce who the spy is.

Our version differs from the original game. In the original one, gameplay occurs in real time, allowing players to initiate votes or accusations at any moment within a time limit. Since we use LLMs as players, real-time play would disadvantage slower models and exacerbate latency issues. We therefore adapt the game to a turn-based format. As a result, the time limit is replaced with a turn limit, which we set to 2 $\times$ the number of players in the game. The overall flow of the game is shown in Figure~\ref{fig:game-flow}, and an example of the running game is presented in Figure~\ref{fig:hook}. 

\subsection{Game Entities $\mathcal{E}$}
In contrast to the original game, we do not limit guessing to places but also include food names as objects to be guessed. We therefore refer to the list of guessable objects as entities, denoted by $\mathcal{E}$. Each game selects one entity from 30 choices. All 30 entities are displayed to each player, with one target entity, $\mathcal{E}_t$, that is not known to the spy. Refer to Appendix~\ref{appendix:entity-selection} for how we define the entities for each language and scenario. 

\subsection{Game Endings} \label{apx: game-end}
The winning conditions for each role are as follows: 

\paragraph{Spy's win conditions} The spy wins if they can deduce $\mathcal{E}_t$ in $\mathcal{P}_{SG}$, a non-spy is majority voted in $\mathcal{P}_{V}$, the turn limit is exceeded, or one of the non-spies quits the game (e.g., a formatting issue in the LLM's output).

\paragraph{Non-spies' win conditions} The non-spy team wins if the spy guesses $\mathcal{E}_t$ incorrectly, the spy gets majority voted in $\mathcal{P}_{V}$, or the spy quits the game (e.g., a formatting issue in the LLM's output).

\subsection{Game Phases $\mathcal{P}$}

We design the game flow to be turn-based, where each turn consists of the following phases: $\mathcal{P}_{QA}$, followed by $\mathcal{P}_{SG}$, and finally $\mathcal{P}_{V}$. We place $\mathcal{P}_{SG}$ before $\mathcal{P}_{V}$ because we prioritize observing the spy’s ability to deduce knowledge, particularly local knowledge.

\paragraph{Question-and-Answer Phase $\mathcal{P}_{QA}$} In this phase, the current player asks another player a question, and the targeted player must answer it. Only one question may be asked per turn. After the turn ends, the answerer becomes the questioner in the next turn.

\paragraph{Spy Guess Phase $\mathcal{P}_{SG}$}
During this phase, the spy may choose to either guess the target entity $\mathcal{E}_t$ or skip the phase. If the spy chooses to guess $\mathcal{E}_t$ and the guess is correct, the spy wins the game; otherwise, the non-spies win.

\paragraph{Voting Phase $\mathcal{P}_{V}$}
In the voting phase, each player votes for another player, accusing them of being the spy. Players may vote for a specific player or choose to skip voting. If a player receives more than half of the votes (e.g., 3 out of 5 players), the game ends, and the accused player's role is revealed. If the accused player is the spy, the non-spies win; otherwise, the spy wins.


\begin{table*}[t]
\centering
\footnotesize
\setlength{\tabcolsep}{2pt}
\renewcommand{\arraystretch}{1.1}
\begin{tabular}{@{}clc|cccc|ccc|ccc@{}}
\toprule
& \textbf{Model} & \textbf{Overall} & \multicolumn{4}{c}{Generic Location} &  \multicolumn{3}{c}{Local Location} &  \multicolumn{3}{c}{Local Food} \\
&  & & G$_{\text{en}}$ & G$_{\text{id}}$ & G$_{\text{arz}}$ & G$_{\text{zh}}$ & L$_{\text{id}}$ & L$_{\text{arz}}$ & L$_{\text{zh}}$ & F$_{\text{id}}$ & F$_{\text{arz}}$ & F$_{\text{zh}}$ \\
\midrule
1 & Gemini-P & \gld{1136}$_{\pm 11}$ & \gld{1086}$_{\pm 33}$ & \gld{1116}$_{\pm 35}$ & \gld{1163}$_{\pm 40}$ & \gld{1174}$_{\pm 39}$ & \gld{1154}$_{\pm 38}$ & \gld{1171}$_{\pm 41}$ & \gld{1161}$_{\pm 42}$ & \slv{1075}$_{\pm 36}$ & \gld{1157}$_{\pm 37}$ & \slv{1136}$_{\pm 38}$ \\
2 & Gemini-F & \slv{1107}$_{\pm 11}$ & \slv{1057}$_{\pm 34}$ & \slv{1071}$_{\pm 34}$ & \slv{1130}$_{\pm 38}$ & \slv{1103}$_{\pm 39}$ & \slv{1086}$_{\pm 34}$ & \slv{1161}$_{\pm 36}$ & \slv{1139}$_{\pm 39}$ & \gld{1098}$_{\pm 35}$ & \slv{1112}$_{\pm 38}$ & \gld{1150}$_{\pm 38}$ \\
3 & Qwen30B-T & \brz{1016}$_{\pm 12}$ & \brz{1045}$_{\pm 33}$ & \fth{982}$_{\pm 35}$ & \fth{1032}$_{\pm 37}$ & \brz{1030}$_{\pm 38}$ & \brz{996}$_{\pm 35}$ & \brz{1071}$_{\pm 36}$ & \fth{985}$_{\pm 38}$ & \fth{1002}$_{\pm 36}$ & \fth{1032}$_{\pm 37}$ & \brz{1005}$_{\pm 34}$ \\
4 & Gemma3-12B & \fth{1003}$_{\pm 11}$ & \fth{995}$_{\pm 32}$ & \brz{1031}$_{\pm 34}$ & \brz{1050}$_{\pm 35}$ & \fth{997}$_{\pm 36}$ & \fth{975}$_{\pm 34}$ & \fth{1000}$_{\pm 38}$ & \brz{994}$_{\pm 36}$ & \brz{1023}$_{\pm 36}$ & 1001$_{\pm 36}$ & \fth{981}$_{\pm 36}$ \\
5 & Qwen3-8B & 967$_{\pm 11}$ & 948$_{\pm 35}$ & 954$_{\pm 33}$ & 964$_{\pm 37}$ & 970$_{\pm 34}$ & 939$_{\pm 33}$ & 959$_{\pm 39}$ & 959$_{\pm 35}$ & 994$_{\pm 33}$ & \brz{1034}$_{\pm 37}$ & 962$_{\pm 35}$ \\
6 & Llama3.1-8B & 771$_{\pm 13}$ & 869$_{\pm 37}$ & 846$_{\pm 39}$ & 660$_{\pm 56}$ & 727$_{\pm 44}$ & 850$_{\pm 37}$ & 638$_{\pm 55}$ & 762$_{\pm 42}$ & 807$_{\pm 41}$ & 664$_{\pm 57}$ & 766$_{\pm 43}$ \\
\bottomrule
\end{tabular}
\caption{\red{Bradley-Terry ratings across scenarios with 95\% bootstrap confidence interval with 1000 resamples. \colorbox{gold!40}{\textbf{Gold}}, \colorbox{silver!50}{Silver}, \colorbox{bronze!30}{Bronze}, and \colorbox{fourth!60}{4th} indicate top 4 per column.}}
\label{tab:bt_elo_rates}
\end{table*}

\subsection{Game Cycles}
In $\mathcal{P}_{QA}$, following the original game rules, the questioner selects a target player, and the target becomes the questioner in the next turn. This allows the same pair of players to target each other in subsequent turns. As a result, other players may have no actions during this phase. 

To address this issue, we introduce two cycles that structure the turns described above: the Round-robin Cycle and the Free Cycle. In the Round-robin Cycle, during $\mathcal{P}_{QA}$, each player must target the next player in a fixed order, ensuring that every player participates at least once as both a questioner and an answerer. During this cycle, the game skips $\mathcal{P}_{SG}$ and $\mathcal{P}_{V}$, as the primary objective is to ensure that all players actively participate in the game.

After all players complete $\mathcal{P}_{QA}$ in the Round-robin Cycle, the game proceeds to the Free Cycle. In this cycle, the default rules apply, including $\mathcal{P}_{QA}$, $\mathcal{P}_{SG}$, and $\mathcal{P}_{V}$.

\subsection{Game Players}

In this setup, both the spy and non-spy roles are played by LLMs. To illustrate, consider a five-player game consisting of four non-spies and one spy. If the non-spies are assigned to Model A and the spy to Model B, each non-spy instance is treated independently despite sharing the same underlying model. We evaluate models across all possible role permutations to ensure a comprehensive all-to-all comparison. Instruction and prompt that each model receives can be seen in Appendix \ref{appendix:prompt}.

\section{Experiment Setup} 

\paragraph{Game Configuration} The games are played with five players, four of whom are non-spies, each played by an LLM independently, and the spy is played by another LLM. The turn limit is five turns in the Round-robin Cycle and five turns in the Free Cycle, totaling 10 turns. The players' order is shuffled to remove positional bias in the analysis. Each game is played once.

\paragraph{Models} We use six models, referred to by the short names in parentheses: \texttt{gemini-2.5-pro} (Gemini-P) and \texttt{gemini-2.5-flash} (Gemini-F)~\cite{comanici2025gemini25pushingfrontier}, \texttt{qwen3-30b-a3b} with thinking enabled (Qwen30B-T) and \texttt{qwen3-8b} (Qwen3-8B)~\cite{yang2025qwen3technicalreport}, \texttt{gemma-3-12b-it} (Gemma3-12B)~\cite{gemmateam2025gemma3technicalreport}, and \texttt{llama-3.1-8b-instruct} (Llama3.1-8B)~\cite{grattafiori2024llama3herdmodels}. We choose LLMs that have multilingual capabilities spanning a range of capabilities, from strong to weak LLMs, so that the ranking leaderboard's pattern can be analyzed. We use OpenRouter to run the inferences. We use a temperature of 1.0. and leave other decoding hyperparameters at their defaults.

\paragraph{Scenarios and Languages} We define three scenarios: Generic (G), where $\mathcal{E}$ consists of the original generic places from the original game, most of which are also used in \citet{kim2025fine}, using the English language and translated to local languages; Local Locations (L), where $\mathcal{E}$ consists of local locations residing in a region that uses the target language; and Local Food (F), where $\mathcal{E}$ consists of food originating from the respective region that uses the target language. We focus on Simplified Chinese (\texttt{zh}), Egyptian Arabic (\texttt{arz}), and Indonesian (\texttt{id}) with their respective countries: China, Egypt, and Indonesia. In the game, each player is shown 30 entities. The list of entities for each scenario can be seen in Appendix~\ref{appendix:entities}.

\paragraph{Evaluation Metrics} To compute the rank of each model, we use the Bradley-Terry Model~\cite{bradley1952rank}, following Chatbot Arena's metric~\cite{chiang2024chatbotarenaopenplatform}, due to its stability in computing the ratings. Additionally, we compare the win rate of each model, where the win rate of a model $M$ is $\frac{\#\text{Wins}}{\#\text{Games Played}_{M}}\times 100$.

We also calculate the \textbf{leakage rate}, which we define as the percentage of games in which a non-spy player reveals the target entity $\mathcal{E}_t$ during $\mathcal{P}_{QA}$. A leakage is considered to have occurred if any non-spy player provides information that explicitly discloses the identity of $\mathcal{E}_t$ to other players, including the spy. The leakage rate of a model $M$ is computed as $\text{Leakage Rate}_{M} = \frac{\text{\#Games with Leakage}_{M}}{\text{\#Non-spy Games}_{M}} \times 100\%$.
In analyzing spy behavior, non-spy behavior, and entity analysis, we use Shannon entropy~\cite{shannon1948mathematical} to calculate how spread out the votes or entity guesses from the players are. 

\section{Overall Game Analysis}

We compare two different settings: overall performance across all languages and scenarios, and the different roles that the model played, with a total of 9,000 matches\red{\footnote{Further statistical robustness analysis and match repetitions can be seen in Appendix~\ref{sec:stat_robust} and Appendix~\ref{apx:cons-check} }}.

\red{
\paragraph{Multicultural Spyfall rankings are consistent with Chatbot Arena on overlapping models} We compare the ranking between our benchmark and Chatbot Arena, accessed on January 1st, 2026. On the five overlapping models\footnote{We omit Qwen3-8B as it is missing from the leaderboard.}, our ranking matches Chatbot Arena's ordering. However, the rating spread differs: the gap between Gemini-P and Gemma3-12B is 133 points in our benchmark (1136 vs 1003) compared to 62 points in Chatbot Arena (1402 vs 1340), and the gap between top and bottom models is much larger in ours. We attribute these differences to several factors. Firstly, our setup samples fewer matches per model (3,000 vs 82,000 for Gemini-P in the Arena). Additionally, Spyfall tests strategic concealment, theory-of-mind reasoning over multi-turn interaction, and culturally specific knowledge, which even the best-performing models struggle with. }

\paragraph{\red{Non-English languages reveal wider performance gaps between models and cause noticeable degradation in weaker models}}
Table~\ref{tab:bt_elo_rates} shows that the ratings of G$_{\text{id}}$, G$_{\text{zh}}$, and G$_{\text{arz}}$ have wider gaps compared to G$_{\text{en}}$. In G$_{\text{en}}$, the second-best model, Gemini-F, and Qwen30B-T have a close gap (12 points), which widens in other scenarios. Additionally, the third and fourth ranks fluctuate between Qwen30B-T and Gemma3-12B. On the other hand, Qwen3-8B and Llama3.1-8B are consistently in fifth and sixth place with a significant gap, with \texttt{arz} being the worst-performing language for Llama3.1-8B.

\paragraph{Local and food scenarios affect performance rankings differently} As shown in Table~\ref{tab:bt_elo_rates}, the rankings in location and food scenarios differ. For instance, in all local location scenarios, Gemini-P consistently outperforms Gemini-F, while in food scenarios, specifically F$_{\text{id}}$ and F$_{\text{zh}}$, Gemini-F outperforms Gemini-P. Additionally, similar to generic locations, the third and fourth places fluctuate between Qwen30B-T and Gemma3-12B in both local location and food scenarios, except in F$_{\text{arz}}$, where Qwen3-8B takes third place. Finally, Llama3.1-8B consistently ranks lowest in all scenarios, with the gap widening in Egyptian Arabic scenarios.

\begin{figure}[htbp]
    \centering
    \includegraphics[width=\linewidth]{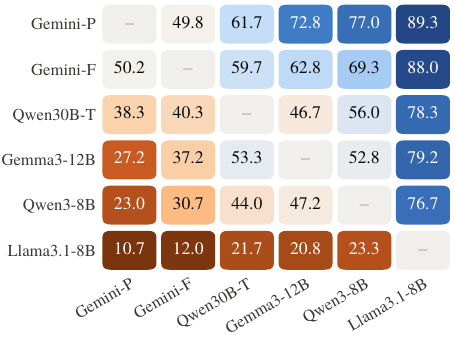}
    \caption{Head-to-head win rate (\%) of the row model against the column model, over 600 matches per pair. Blue: the row model wins more often; orange: it loses more often.}
    \label{fig:win_rate_comparisons}
\end{figure}

\paragraph{Some models perform better against specific opponents} From the match matrices in Figure~\ref{fig:win_rate_comparisons}, all models have a win rate of more than 75\% against Llama3.1-8B, demonstrating an easy matchup that explains the low rating of Llama3.1-8B across all scenarios. Gemini-P and Gemini-F have a close matchup (approximately 50\% win rate) against each other, indicating a competitive pairing. Despite this closeness, when models other than Gemini-F face Gemini-P, their win rates are lower than when facing Gemini-F, demonstrating Gemini-P's stronger performance against other models compared to Gemini-F. Interestingly, Qwen30B-T has less than a 47\% win rate against Gemma3-12B despite having a higher rating, indicating an unfavorable matchup for Qwen30B-T against Gemma3-12B. However, Gemma3-12B has a lower win rate against Gemini-P and Gemini-F compared to Qwen30B-T.

\section{What Skills Can Multicultural Spyfall Show?}

Table~\ref{tab:bt_elo_rates} reports ratings computed from win and loss outcomes, which capture relative performance between models. If a scenario makes all models perform worse by a similar amount, their pairwise win rates against each other barely change, and the rating gaps remain stable across scenarios and languages (See Appendix~\ref{apx:why-similar} for detailed observation). Table~\ref{tab:bt_elo_rates} is therefore useful for comparing models overall, but it does not reveal the skills that the models have. To succeed in this game, we argue that a model at least needs these capabilities: language proficiency (producing fluent, rule-compliant output in the target language), cultural knowledge (recognizing and reasoning about region-specific entities), and strategic reasoning (modeling other players' beliefs to conceal or detect roles). In this section, we observe the models by categorizing the analysis along these three axes to elucidate each model's skills.

\subsection{Language Proficiency}


\begin{table}[t!]
    \centering
    \small
    \setlength{\tabcolsep}{5pt}
    \begin{tabular}{lrrrr}
        \toprule
        Model & \texttt{en} & \texttt{id} & \texttt{zh} & \texttt{arz} \\
        \midrule
        Gemini-F & 0.0 & 0.0 & 0.0 & 0.0 \\
        Gemini-P & 0.0 & 0.0 & 0.0 & 0.0 \\
        Gemma3-12B & 0.0 & 1.6 & 4.9 & 1.6 \\
        Llama3.1-8B & 17.3 & 37.1 & 29.6 & 13.6 \\
        Qwen30B-T & 0.7 & 0.4 & 0.0 & 0.0 \\
        Qwen3-8B & 0.0 & 1.8 & 8.7 & 2.4 \\
        \bottomrule
    \end{tabular}
    \caption{Non-Spy Information Leakage Rates (\%) by Language. }
    \label{tab:leakage_by_lang}
\end{table}

\paragraph{Llama3.1-8B Has High Leakage Rates Up to 37\%, While Gemini Models Have None} In this experiment, despite being informed about the game rules, some models leak the $\mathcal{E}_t$ directly, which may allow the spy to guess it easily. Table~\ref{tab:leakage_by_lang} shows the leakage rate across models and languages. Llama3.1-8B has the highest leakage rate compared to other models, particularly in \texttt{id}, reaching 37\%. In contrast, Gemini-F and Gemini-P have 0\% leakage across all languages. Interestingly, Qwen3-8B has a moderate leakage rate in \texttt{zh} ($\sim$9\%), despite being heavily trained on \texttt{zh} data. This suggests that language capability does not directly correlate with the leakage rate.

\begin{table}[t]
\centering
\small
\setlength{\tabcolsep}{4pt}
\begin{tabular}{lccc}
\toprule
\textbf{Model} & \textbf{id} & \textbf{zh} & \textbf{arz} \\
\midrule
Gemini-F & \makecell{\texttt{id} 98.5\\\texttt{ms} 1.0} & \makecell{\texttt{zh} 99.9\\\texttt{wuu} 0.1} & \makecell{\texttt{arz} 70.89\\\texttt{apd} 10.45} \\
\midrule
Gemini-P & \makecell{\texttt{id} 98.7\\\texttt{ms} 1.2} & \makecell{\texttt{zh} 99.9\\\texttt{ja} 0.1} & \makecell{\texttt{arz} 80.47\\\texttt{apd} 7.56} \\
\midrule
Gemma3-12B   & \makecell{\texttt{id} 98.8\\\texttt{ms} 0.9} & \makecell{\texttt{zh} 99.5\\\texttt{en} 0.2} & \makecell{\texttt{msa} 59.44\\\texttt{acx} 19.89} \\
\midrule
Llama3.1-8B & \makecell{\texttt{id} 96.6\\\texttt{ms} 2.3} & \makecell{\texttt{zh} 89.2\\\texttt{en} 8.0} & \makecell{\texttt{msa} 29.76\\\texttt{acx} 28.99} \\
\midrule
Qwen30B-T & \makecell{\texttt{id} 98.9\\\texttt{ms} 0.7} & \makecell{\texttt{zh} 95.4\\\texttt{en} 4.2} & \makecell{\texttt{msa} 49.51\\\texttt{acx} 25.93} \\
\midrule
Qwen3-8B & \makecell{\texttt{id} 98.6\\\texttt{ms} 0.6} & \makecell{\texttt{zh} 99.9\\\texttt{en} 0.1} & \makecell{\texttt{msa} 50.3\\\texttt{acx} 34.70} \\
\bottomrule
\end{tabular}
\caption{\red{Language detection of outputs. Top-1 (first) / Top-2 (second).}}
\label{tab:lang_dist}
\end{table}

\red{\paragraph{Models Follow Target Languages but Struggle With a Dialect} Table~\ref{tab:lang_dist} shows the language used by each model across all scenarios. We identified the language using fastText~\citet{joulin-etal-2017-bag}. Then, we processed rows detected as Arabic with Arabic dialect identification trained on the MADAR Corpus~\cite{inoue-etal-2021-interplay}, which has demonstrated better performance on fine-grained Arabic dialect distinctions.
Overall, most models are consistent in using the target language, with more than 95\% usage (except for Llama3.1-8B model) with the exception with Arabic dialect. These results show that Qwen3-8B, Qwen30B-T, Llama3.1-8B, and Gemma3-12B predominantly generate Modern Standard Arabic (msa) rather than the target arz, with \texttt{arz} usage below 13\% across these models. Even Gemini-P and Gemini-F, which show substantially higher \texttt{arz} usage (\textasciitilde70-80\%), occasionally produce Sudanese Arabic (apd), a geographically close and linguistically overlapping dialect~\cite{keleg-etal-2025-revisiting}. Supported by our manual check by sampling some of the games, most models struggle to consistently adhere to  the\texttt{arz} dialect despite explicit instructions, instead defaulting to \texttt{msa} which follows behavior observed in~\citet{robinson-etal-2025-al}.}
\begin{table*}[htbp]
\centering

\small
\begin{tabular}{lp{3cm}ccp{8cm}}
\hline
\textbf{Lang} & \textbf{$\mathcal{E}_t$} & \textbf{H} & \textbf{Acc} & \textbf{Guess Distribution (Top-5)} \\
\hline
id & Loloh Cemcem & 3.21 & 9\% & \textit{Es Lidah Buaya} (4); \textit{Martabak} (4); \textit{Bandrek} (3); \textit{Beras Kencur} (3); \checkmark\textit{Loloh Cemcem} (2) \\
arz & \arabictext{ترمس} (Turmus) & 3.43 & 5\% & \textit{\arabictext{حمص الشام}} (Hummus al-Sham) (3); \textit{\arabictext{كحك العيد}} (Kahk el-Eid) (3); \textit{\arabictext{حواوشي}} (Hawawshi) (2); \textit{\arabictext{سحلب}} (Sahlab) (2); \textit{\arabictext{كشري}} (Koshary) (2) \\
zh & \zh{馕} (Naan) & 2.49 & 29\% & \textit{\zh{烧饼} (Shaobing)} (5); \checkmark \textit{\zh{馕} (Naan)} (5); \textit{\zh{元宵} (Yuanxiao)} (2); \textit{\zh{肉夹馍} (Roujiamo)} (2); \textit{\zh{手抓饼} (Shouzhuabing)} (1) \\
\hline
\multicolumn{5}{l}{\small H = Entropy (bits); Acc = Accuracy (\%); \checkmark = Correct guess} \\
\end{tabular}
\caption{Detailed Guess Distribution for High-Entropy Locations and low accuracy $\mathcal{E}_t$}
\label{tab:high_entropy_detailed}
\end{table*}

\begin{table}[t]
\centering
\small
\setlength{\tabcolsep}{3pt}
\begin{tabular}{lcccccc}
\toprule
& \multicolumn{2}{c}{\textbf{G}} & \multicolumn{2}{c}{\textbf{L}} & \multicolumn{2}{c}{\textbf{F}} \\
\cmidrule(lr){2-3} \cmidrule(lr){4-5} \cmidrule(lr){6-7}
& Acc & H & Acc & H & Acc & H \\
\midrule
en & 67.3 & 1.49 & -- & -- & -- & -- \\
id & 67.1 & 1.49 & 49.6 & 1.81 & 40.9 & 2.35 \\
zh & 64.2 & 1.55 & 54.5 & 1.50 & 51.6 & 2.14 \\
arz & 59.5 & 1.69 & 33.8 & 2.10 & 24.4 & 2.80 \\
\bottomrule
\end{tabular}
\caption{Overall Accuracy (Acc) and Entropy (H) by Dataset}
\label{tab:dataset_acc_entropy}
\end{table}

\begin{figure}[htbp]
    \centering
    \includegraphics[width=\linewidth]{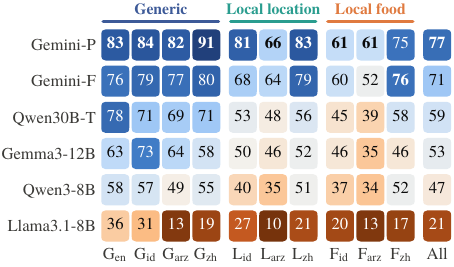}
    \caption{Spy win rate (\%) by model and scenario. Colors are centered at 50\% (blue: the spy wins more often than it loses; orange: less often). \textbf{Bold}: best model per column; \textit{All} pools every scenario.}
    \label{fig:win_rate_scenario}
\end{figure}

\paragraph{Some models have an advantage in certain languages where they gain higher spy win rates while others decline in Generic scenario} As can be seen in Figure~\ref{fig:win_rate_scenario}, compared to G$_{\text{en}}$, Gemini family models improve their spy win rate by up to 8\% in G$_{\text{id}}$ and G$_{\text{zh}}$, while Qwen30B-T and Qwen3-8B show moderate drops (up to 7\% and 3\%, respectively) in spy win rate, followed by a huge drop for Llama3.1-8B in G$_{\text{zh}}$ ($\sim$17\%). Meanwhile, in G$_{\text{arz}}$, the Qwen and Llama models significantly tend to drop the most in spy win rate, showing that \texttt{arz} is more challenging for these models, while others have slight changes.

\paragraph{Qwen3-8B and Qwen30B-T have the highest spy win rates in F and L scenarios in \texttt{zh} language compared to other languages} In Figure~\ref{fig:win_rate_scenario}, both models achieve their highest spy win rates in \texttt{zh}. For F, the gap is 13--15 percentage points over \texttt{id} (Qwen30B-T: 58\% vs.\ 45\%; Qwen3-8B: 52\% vs.\ 37\%) and 18--19 points over \texttt{arz} (39\%/34\%). The same trend holds for L, though with smaller margins (Qwen30B-T: 56\% vs.\ 53\%/48\%; Qwen3-8B: 51\% vs.\ 40\%/35\%). This suggests that these models have better local knowledge of Chinese food and, to a lesser extent, locations, which is expected as they are heavily trained on \texttt{zh} data.

\subsection{Cultural Knowledge} \label{sec:cultural-knowledge}

\paragraph{Each language scenario differs in difficulty for guessing}
\red{
Egyptian Arabic (\texttt{arz}) consistently proves most difficult for spies to guess correctly across all scenarios (Table~\ref{tab:dataset_acc_entropy}). In generic locations, spy guess accuracy is: \texttt{en} (67.3\%), \texttt{id} (67.1\%), \texttt{zh} (64.2\%), and \texttt{arz} (59.5\%). This gap widens dramatically in local contexts, where for local food, accuracy drops to: \texttt{zh} (51.6\%), \texttt{id} (40.9\%), and \texttt{arz} (24.4\%), representing a more than $2\times$ difference between the easiest and hardest. Local locations show a similar pattern: \texttt{zh} (54.5\%), \texttt{id} (49.6\%), and \texttt{arz} (33.8\%).
}

\paragraph{Overall, location and food scenarios in spy role have lower win rates compared to generic location} In Figure~\ref{fig:win_rate_scenario}, in most models, the spy win rate in local location and food scenarios drops significantly compared to the generic location scenario, indicating that predicting local knowledge is more challenging for the spy role. Similar to the generic location scenario, the \texttt{arz} language has the lowest spy win rate compared to other languages. Both location and food scenarios have similar spy win rates across models.

\paragraph{Local Scenarios and Food Are More Challenging than Generic Locations} 
We evaluate the difficulty of each scenario by analyzing the spy players' ability to guess $\mathcal{E}_t$, measured through overall accuracy and guess entropy across entities (Table~\ref{tab:dataset_acc_entropy}). 
Overall, generic scenarios exhibit higher guess accuracy and lower guess entropy compared to local locations and food scenarios, indicating that generic locations are easier for the spy to deduce. 
This performance gap likely stems from the fact that generic locations are more globally recognized by models, whereas local entities require specialized regional knowledge. 
Notably, in both local location and food scenarios, the \texttt{arz} language exhibits the lowest overall entity accuracy and the highest guess entropy compared to other languages.

\paragraph{Lower-Accuracy Entities Share Characteristics with Distractors} 

We sampled local food entities with the lowest accuracy to identify common patterns of confusion, as shown in Table~\ref{tab:high_entropy_detailed}.
In \texttt{id}, the entity \textit{Loloh Cemcem} refers to a traditional drink; it was frequently confused with similar beverages such as \textit{Es Lidah Buaya}, \textit{Bandrek}, and \textit{Beras Kencur}. 
In \texttt{zh}, \zh{馕} (Naan) refers to a staple flatbread that shares significant characteristics with \zh{烧饼} (Shaobing), \zh{肉夹馍} (Roujiamo), and \zh{手抓饼} (Shouzhua Bing). 
In \texttt{arz}, \arabictext{ترمس} (Turmus, boiled lupin beans) is a staple street snack in Egypt. While it shares characteristics with other snacks, these foods are distinct; however, the models fail to take this into consideration.

\paragraph{Some Errors are due to Model's Lack of Cultural Knowledge} 
As shown in Table~\ref{tab:high_entropy_detailed}, some confusion occurs between entirely unrelated entities. For example, the Indonesian traditional healthy drink was quite often confused with \textit{Martabak}, a pancake-like street food. Upon further investigation, we note that some weaker non-spies did not know the drink and answered incorrectly, as if it were a fried food. The spy, thinking it was a dessert, also guessed incorrectly. Stronger models such as Gemini tend to discuss the factual information regarding the drink correctly.

\subsection{Strategic Reasoning}

\begin{table}[htbp]
\centering
\small
\begin{tabular}{lrr}
\toprule
\textbf{Category} & \textbf{Count} & \textbf{\%} \\
\midrule
Spy Guess Wrong & 2,917 & 32.41 \\
Spy Guess Correct & 3,257 & 36.19 \\
Vote Majority to Spy & 716 & 7.96 \\
Vote Majority to Non-Spy & 363 & 4.03 \\
Spy Quit & 458 & 5.09 \\
Non-Spy Quit & 1,289 & 14.32 \\
\bottomrule
\end{tabular}
\caption{Overall Game Outcome Statistics}
\label{tab:overall_action_stat}
\end{table}

\paragraph{Game Endings Are Dominated by Spy Guess Actions} Table~\ref{tab:overall_action_stat} shows the overall game outcome statistics across all models and scenarios. The most frequent game ending is when the spy guesses the entity, comprising 36.19\% correct guesses and 32.41\% wrong guesses, summing to 68.6\%. This shows that the spy is more likely to win through guessing rather than being voted out. Additionally, quit actions are also significant ($\approx$19.41\%), dominated by the Llama3.1-8B model (70.52\% of all quits). Finally, the voting phase only contributes to 12\% of all game endings, showing that the models are less likely to vote out the spy.

\paragraph{Why Does a Spy Player Get Voted Out?} 
Through a qualitative analysis of 20 randomly sampled matches where Gemini-P or Flash were eliminated, we identified three primary reasons for their failure despite their underlying strength. 
First and most prevalent, spies often gave \textbf{generic or incorrect answers} while non-spies provided specific signals about the target entity ($\mathcal{E}_t$). 
For instance, in an L$_{\text{id}}$ match where $\mathcal{E}_t$ is ``Binus,'' the spy responded vaguely about costs (``it depends on perspective...'' translated from \texttt{id}), failing to recognize it as a famously expensive private university, which effectively revealed their identity. 
Second, models occasionally asked \textbf{``obvious fishing questions''} to gather information. In one L$_{\text{zh}}$ match involving ``Peking University,'' the spy asked about the ``main users'' of the place, even though the academic context had already been explicitly established by non-spies. 
Finally, matches in L$_{\text{arz}}$ and L$_{\text{zh}}$ showed spies being \textbf{voted out for targeting innocent non-spies}, triggering retaliatory votes that eliminated the spy regardless of their previous answer quality.

\begin{table}[t]
\centering
\small
\begin{tabular}{l|ccc}
\toprule
\textbf{Model} & \textbf{Guess $\times$} & \textbf{Got Voted} & \textbf{Quit} \\
\midrule
Llama3.1-8B & 82.4 & 5.4 & 12.2 \\
Qwen3-8B & 81.1 & 7.1 & 11.8 \\
Qwen30B-T & 80.8 & 6.6 & 12.6 \\
Gemma3-12B & 78.0 & 8.8 & 13.2 \\
Gemini-P & 63.6 & 25.9 & 10.5 \\
Gemini-F & 60.8 & 29.4 & 9.7 \\
\midrule
\textbf{Average} & \textbf{74.5} & \textbf{13.9} & \textbf{11.7} \\
\bottomrule
\end{tabular}
\caption{Non-Spy Victory Distribution by Model (when model is non-spy). Column denotes spy last action. Percentages show the win rate win methods of Non-spies}
\label{tab:nonspy_win_methods}
\end{table}

\begin{figure}[htbp]
    \centering
    \includegraphics[width=\linewidth]{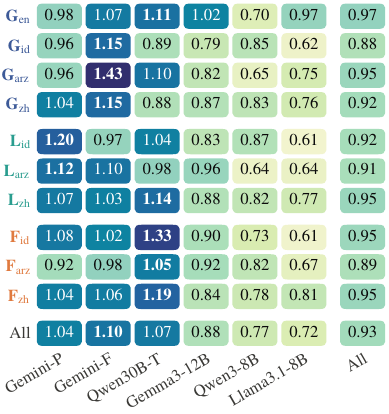}
    \caption{Spy vote dispersion score by scenario (rows) and model playing the spy (columns); higher is better for the spy. \textbf{Bold}: best model per row; \textit{All} is the unweighted mean over models (column) or scenarios (row).}
    \label{fig:spy_vote_dispersion}
\end{figure}

\paragraph{Spy Vote Dispersion Varies Depends on The Model Capacity} in In calculating the voting session when the spy is present, we introduce \textbf{Vote Dispersion}, which can be calculated using $H \times (1-V_S)$, where $H$ is the Shannon entropy value and $V_S$ denotes the percentage of votes that the spy receives (1 means everyone votes for the spy) in a single voting session. This equation penalizes cases where entropy is low but the votes target the spy. A higher value is better for the spy player. The visualization (Figure~\ref{fig:spy_vote_dispersion}) reveals the vote dispersion, where Gemini-F (1.095), Qwen30B-T (1.070), and Gemini-P (1.037) achieve the highest dispersion scores, indicating superior ability to manipulate voting patterns and evade detection when acting as the spy. In contrast, smaller models like Llama3.1-8B (0.721) and Qwen3-8B (0.770) struggle significantly, being more easily identified and voted out. Scenario differences are relatively minor (0.878 to 0.974), suggesting that spy evasion success depends more on model capability than on language or local domain. The overall average of 0.929 indicates moderate voting chaos across all conditions.



More analysis on probing this section can be seen further in Appendix~\ref{apx:add-behv}
\section{Related Works}

\paragraph{Dynamic and Game-Based LLM Evaluation} Recent research has increasingly leveraged strategic games to evaluate LLMs' capabilities. Some benchmarks assess LLMs capability on games, such as
social deduction games. For instance, AvalonBench \cite{light2023avalonbenchevaluatingllmsplaying} evaluates deception and negotiation skills, while several studies utilize Werewolf to assess emergent strategic behaviors \cite{bailis2024werewolfarenacasestudy, agarwal2025wolfwerewolfbasedobservationsllm, song2025survivalevaluatingllmssocial}. Additionally, AmongAgents \cite{chi2024amongagentsevaluatinglargelanguage} adapts the game Among Us into a fully text-based format. Similarly, \cite{kim2025fine} investigates Spyfall, though their evaluation is limited to generic settings in English.

Beyond social deduction, debate-based evaluation \cite{moniri-etal-2025-debate} and interactive fiction \cite{hausknecht2020interactivefictiongamescolossal, cote2019textworldlearningenvironmenttextbased} probe oversight and reasoning capabilities. Strategic benchmarks like GTBench \cite{duan2024gtbenchuncoveringstrategicreasoning} and GameBench \cite{costarelli2024gamebench} reveal that LLMs excel in probabilistic scenarios but struggle with complete-information games. Some works also collect aggregated multi-agent benchmarks like AgentBench \cite{liu2025agentbenchevaluatingllmsagents}.

\paragraph{Multilingual and Multicultural Benchmark} Culture is often operationalized through proxies representing specific concepts~\cite{adilazuarda-etal-2024-towards}. Several works have constructed static multicultural benchmarks within native language contexts such as CulturalBench~\cite{chiu-etal-2025-culturalbench}, BLEnD~\cite{myung2025blendbenchmarkllmseveryday}, GlobalPIQA~\cite{chang2025global}, ArabCulture~\cite{sadallah-etal-2025-commonsense}, or COPAL-ID~\cite{wibowo-etal-2024-copal}. These works mostly show that models struggle with non-western contexts. However, these benchmarks typically rely on static formats that are susceptible to data contamination and performance saturation, limitations that we address in this paper.

\section{Conclusion}

We introduce a dynamic benchmarking framework based on Spyfall to evaluate LLMs on their multilingual and multicultural reasoning. Our results show that while top models perform well, there is still a noticeable drop in capability when models are tested in non-English languages or on specific cultural topics.  The framework also reveals several distinct behaviors in the game across different aspects. For instance, in the language aspect, some models struggle to adhere \texttt{to the arz} language. In the cultural context, models lack cultural understanding, especially when similarities with some entities in the pool exist. Finally, in the strategic aspect, some models are able to misdirect other models, for instance, exhibited by the Gemini  families. This framework offers a reliable and scalable evaluation tool for how well models truly understand the diverse local cultural contexts.

Future work will explore the application of dynamic benchmarking to interactive tasks like debate to further assess multilingual and multicultural nuances, as our framework is significantly more scalable and offers greater resilience to data leakage and saturation.

\section*{Limitations}

While this study focuses on three countries and their primary languages, it establishes a robust framework for evaluating regional linguistic nuances. Although the current dataset includes 30 local landmarks and 30 traditional foods per country, the benchmark is designed to be highly scalable. Its dynamic architecture allows for the seamless integration of additional cultural entities and languages in future iterations.

Finally, employing four identical model instances for the non-spy team may introduce behavioral coupling. However, the modular nature of our environment fully supports the future exploration of heterogeneous, many-to-many model interactions.

\section*{Ethical Considerations}

\textbf{Cultural Representation and Bias:} While we aim to evaluate multicultural understanding, we acknowledge that our selection of 30 food items and 30 landmarks per country is non-exhaustive. These entities represent a subset of cultural identities and may reflect certain regional biases within the chosen countries. We have made efforts to include widely recognized cultural markers to ensure fair testing.

\textbf{Intentional Deception:} The social deduction framework involves models playing the role of a ``Spy,'' which requires the use of strategic concealment or misdirection. We emphasize that this is used strictly as a proxy for reasoning and communication capability within a game context and should not be applied to encourage harmful deceptive behavior in real-world applications.

\textbf{LLM Usage:} We use LLMs such as Gemini and Writefull (Overleaf) to correct the grammar in our writing. 

\bibliography{custom}

@inproceedings{keleg-etal-2025-revisiting,
    title = "Revisiting Common Assumptions about {A}rabic Dialects in {NLP}",
    author = "Keleg, Amr  and
      Goldwater, Sharon  and
      Magdy, Walid",
    editor = "Che, Wanxiang  and
      Nabende, Joyce  and
      Shutova, Ekaterina  and
      Pilehvar, Mohammad Taher",
    booktitle = "Proceedings of the 63rd Annual Meeting of the Association for Computational Linguistics (Volume 1: Long Papers)",
    month = jul,
    year = "2025",
    address = "Vienna, Austria",
    publisher = "Association for Computational Linguistics",
    url = "https://aclanthology.org/2025.acl-long.166/",
    doi = "10.18653/v1/2025.acl-long.166",
    pages = "3309--3327",
    ISBN = "979-8-89176-251-0"
}

@inproceedings{inoue-etal-2021-interplay,
    title = "The Interplay of Variant, Size, and Task Type in {A}rabic Pre-trained Language Models",
    author = "Inoue, Go  and
      Alhafni, Bashar  and
      Baimukan, Nurpeiis  and
      Bouamor, Houda  and
      Habash, Nizar",
    editor = "Habash, Nizar  and
      Bouamor, Houda  and
      Hajj, Hazem  and
      Magdy, Walid  and
      Zaghouani, Wajdi  and
      Bougares, Fethi  and
      Tomeh, Nadi  and
      Abu Farha, Ibrahim  and
      Touileb, Samia",
    booktitle = "Proceedings of the Sixth Arabic Natural Language Processing Workshop",
    month = apr,
    year = "2021",
    address = "Kyiv, Ukraine (Virtual)",
    publisher = "Association for Computational Linguistics",
    url = "https://aclanthology.org/2021.wanlp-1.10/",
    pages = "92--104"
}

@misc{chiang2024chatbotarenaopenplatform,
  title         = {Chatbot Arena: An Open Platform for Evaluating LLMs by Human Preference},
  author        = {Wei-Lin Chiang and Lianmin Zheng and Ying Sheng and Anastasios Nikolas Angelopoulos and Tianle Li and Dacheng Li and Hao Zhang and Banghua Zhu and Michael Jordan and Joseph E. Gonzalez and Ion Stoica},
  year          = {2024},
  eprint        = {2403.04132},
  archiveprefix = {arXiv},
  primaryclass  = {cs.AI},
  url           = {https://arxiv.org/abs/2403.04132}
}

@article{bradley1952rank,
  title     = {Rank analysis of incomplete block designs: I. the method of paired comparisons},
  author    = {Bradley, Ralph Allan and Terry, Milton E},
  journal   = {Biometrika},
  volume    = {39},
  number    = {3/4},
  pages     = {324--345},
  year      = {1952},
  publisher = {JSTOR}
}

@article{shannon1948mathematical,
  title={A Mathematical Theory of Communication},
  author={Shannon, Claude E.},
  journal={The Bell System Technical Journal},
  volume={27},
  number={3},
  pages={379--423},
  year={1948},
  publisher={Nokia Bell Labs}
}

@inproceedings{joulin-etal-2017-bag,
    title = "Bag of Tricks for Efficient Text Classification",
    author = "Joulin, Armand  and
      Grave, Edouard  and
      Bojanowski, Piotr  and
      Mikolov, Tomas",
    editor = "Lapata, Mirella  and
      Blunsom, Phil  and
      Koller, Alexander",
    booktitle = "Proceedings of the 15th Conference of the {E}uropean Chapter of the Association for Computational Linguistics: Volume 2, Short Papers",
    month = apr,
    year = "2017",
    address = "Valencia, Spain",
    publisher = "Association for Computational Linguistics",
    url = "https://aclanthology.org/E17-2068/",
    pages = "427--431"
}

@misc{light2023avalonbenchevaluatingllmsplaying,
      title={AvalonBench: Evaluating LLMs Playing the Game of Avalon}, 
      author={Jonathan Light and Min Cai and Sheng Shen and Ziniu Hu},
      year={2023},
      eprint={2310.05036},
      archivePrefix={arXiv},
      primaryClass={cs.AI},
      url={https://arxiv.org/abs/2310.05036}, 
}

@misc{chi2024amongagentsevaluatinglargelanguage,
      title={AMONGAGENTS: Evaluating Large Language Models in the Interactive Text-Based Social Deduction Game}, 
      author={Yizhou Chi and Lingjun Mao and Zineng Tang},
      year={2024},
      eprint={2407.16521},
      archivePrefix={arXiv},
      primaryClass={cs.CL},
      url={https://arxiv.org/abs/2407.16521}, 
}

@inproceedings{wibowo-etal-2024-copal,
    title = "{COPAL}-{ID}: {I}ndonesian Language Reasoning with Local Culture and Nuances",
    author = "Wibowo, Haryo  and
      Fuadi, Erland  and
      Nityasya, Made  and
      Prasojo, Radityo Eko  and
      Aji, Alham",
    editor = "Duh, Kevin  and
      Gomez, Helena  and
      Bethard, Steven",
    booktitle = "Proceedings of the 2024 Conference of the North American Chapter of the Association for Computational Linguistics: Human Language Technologies (Volume 1: Long Papers)",
    month = jun,
    year = "2024",
    address = "Mexico City, Mexico",
    publisher = "Association for Computational Linguistics",
    url = "https://aclanthology.org/2024.naacl-long.77/",
    doi = "10.18653/v1/2024.naacl-long.77",
    pages = "1404--1422"
}

@inproceedings{sadallah-etal-2025-commonsense,
    title = "Commonsense Reasoning in {A}rab Culture",
    author = "Sadallah, Abdelrahman  and
      Tonga, Junior Cedric  and
      Almubarak, Khalid  and
      Almheiri, Saeed  and
      Atif, Farah  and
      Qwaider, Chatrine  and
      Kadaoui, Karima  and
      Shatnawi, Sara  and
      Alesh, Yaser  and
      Koto, Fajri",
    editor = "Che, Wanxiang  and
      Nabende, Joyce  and
      Shutova, Ekaterina  and
      Pilehvar, Mohammad Taher",
    booktitle = "Proceedings of the 63rd Annual Meeting of the Association for Computational Linguistics (Volume 1: Long Papers)",
    month = jul,
    year = "2025",
    address = "Vienna, Austria",
    publisher = "Association for Computational Linguistics",
    url = "https://aclanthology.org/2025.acl-long.380/",
    doi = "10.18653/v1/2025.acl-long.380",
    pages = "7695--7710",
    ISBN = "979-8-89176-251-0"
}

@article{chang2025global,
  title={Global PIQA: Evaluating Physical Commonsense Reasoning Across 100+ Languages and Cultures},
  author={Chang, Tyler A and Arnett, Catherine and Eldesokey, Abdelrahman and Sadallah, Abdelrahman and Kashar, Abeer and Daud, Abolade and Olanihun, Abosede Grace and Mohammed, Adamu Labaran and Praise, Adeyemi and Sharma, Adhikarinayum Meerajita and others},
  journal={arXiv preprint arXiv:2510.24081},
  year={2025}
}

@misc{liu2025agentbenchevaluatingllmsagents,
      title={AgentBench: Evaluating LLMs as Agents}, 
      author={Xiao Liu and Hao Yu and Hanchen Zhang and Yifan Xu and Xuanyu Lei and Hanyu Lai and Yu Gu and Hangliang Ding and Kaiwen Men and Kejuan Yang and Shudan Zhang and Xiang Deng and Aohan Zeng and Zhengxiao Du and Chenhui Zhang and Sheng Shen and Tianjun Zhang and Yu Su and Huan Sun and Minlie Huang and Yuxiao Dong and Jie Tang},
      year={2025},
      eprint={2308.03688},
      archivePrefix={arXiv},
      primaryClass={cs.AI},
      url={https://arxiv.org/abs/2308.03688}, 
}

@article{costarelli2024gamebench,
  title={Gamebench: Evaluating strategic reasoning abilities of llm agents},
  author={Costarelli, Anthony and Allen, Mat and Hauksson, Roman and Sodunke, Grace and Hariharan, Suhas and Cheng, Carlson and Li, Wenjie and Clymer, Joshua and Yadav, Arjun},
  journal={arXiv preprint arXiv:2406.06613},
  year={2024}
}

@misc{duan2024gtbenchuncoveringstrategicreasoning,
      title={GTBench: Uncovering the Strategic Reasoning Limitations of LLMs via Game-Theoretic Evaluations}, 
      author={Jinhao Duan and Renming Zhang and James Diffenderfer and Bhavya Kailkhura and Lichao Sun and Elias Stengel-Eskin and Mohit Bansal and Tianlong Chen and Kaidi Xu},
      year={2024},
      eprint={2402.12348},
      archivePrefix={arXiv},
      primaryClass={cs.CL},
      url={https://arxiv.org/abs/2402.12348}, 
}

@misc{cote2019textworldlearningenvironmenttextbased,
      title={TextWorld: A Learning Environment for Text-based Games}, 
      author={Marc-Alexandre Côté and Ákos Kádár and Xingdi Yuan and Ben Kybartas and Tavian Barnes and Emery Fine and James Moore and Ruo Yu Tao and Matthew Hausknecht and Layla El Asri and Mahmoud Adada and Wendy Tay and Adam Trischler},
      year={2019},
      eprint={1806.11532},
      archivePrefix={arXiv},
      primaryClass={cs.LG},
      url={https://arxiv.org/abs/1806.11532}, 
}

@misc{hausknecht2020interactivefictiongamescolossal,
      title={Interactive Fiction Games: A Colossal Adventure}, 
      author={Matthew Hausknecht and Prithviraj Ammanabrolu and Marc-Alexandre Côté and Xingdi Yuan},
      year={2020},
      eprint={1909.05398},
      archivePrefix={arXiv},
      primaryClass={cs.AI},
      url={https://arxiv.org/abs/1909.05398}, 
}

@misc{myung2025blendbenchmarkllmseveryday,
      title={BLEnD: A Benchmark for LLMs on Everyday Knowledge in Diverse Cultures and Languages}, 
      author={Junho Myung and Nayeon Lee and Yi Zhou and Jiho Jin and Rifki Afina Putri and Dimosthenis Antypas and Hsuvas Borkakoty and Eunsu Kim and Carla Perez-Almendros and Abinew Ali Ayele and Víctor Gutiérrez-Basulto and Yazmín Ibáñez-García and Hwaran Lee and Shamsuddeen Hassan Muhammad and Kiwoong Park and Anar Sabuhi Rzayev and Nina White and Seid Muhie Yimam and Mohammad Taher Pilehvar and Nedjma Ousidhoum and Jose Camacho-Collados and Alice Oh},
      year={2025},
      eprint={2406.09948},
      archivePrefix={arXiv},
      primaryClass={cs.CL},
      url={https://arxiv.org/abs/2406.09948}, 
}

@misc{agarwal2025wolfwerewolfbasedobservationsllm,
      title={WOLF: Werewolf-based Observations for LLM Deception and Falsehoods}, 
      author={Mrinal Agarwal and Saad Rana and Theo Sundoro and Hermela Berhe and Spencer Kim and Vasu Sharma and Sean O'Brien and Kevin Zhu},
      year={2025},
      eprint={2512.09187},
      archivePrefix={arXiv},
      primaryClass={cs.MA},
      url={https://arxiv.org/abs/2512.09187}, 
}

@misc{bailis2024werewolfarenacasestudy,
      title={Werewolf Arena: A Case Study in LLM Evaluation via Social Deduction}, 
      author={Suma Bailis and Jane Friedhoff and Feiyang Chen},
      year={2024},
      eprint={2407.13943},
      archivePrefix={arXiv},
      primaryClass={cs.CL},
      url={https://arxiv.org/abs/2407.13943}, 
}

@inproceedings{adilazuarda-etal-2024-towards,
    title = "Towards Measuring and Modeling ``Culture'' in {LLM}s: A Survey",
    author = "Adilazuarda, Muhammad Farid  and
      Mukherjee, Sagnik  and
      Lavania, Pradhyumna  and
      Singh, Siddhant Shivdutt  and
      Aji, Alham Fikri  and
      O{'}Neill, Jacki  and
      Modi, Ashutosh  and
      Choudhury, Monojit",
    editor = "Al-Onaizan, Yaser  and
      Bansal, Mohit  and
      Chen, Yun-Nung",
    booktitle = "Proceedings of the 2024 Conference on Empirical Methods in Natural Language Processing",
    month = nov,
    year = "2024",
    address = "Miami, Florida, USA",
    publisher = "Association for Computational Linguistics",
    url = "https://aclanthology.org/2024.emnlp-main.882/",
    doi = "10.18653/v1/2024.emnlp-main.882",
    pages = "15763--15784"
}

@inproceedings{chiu-etal-2025-culturalbench,
    title = "{C}ultural{B}ench: A Robust, Diverse and Challenging Benchmark for Measuring {LM}s' Cultural Knowledge Through Human-{AI} Red-Teaming",
    author = "Chiu, Yu Ying  and
      Jiang, Liwei  and
      Lin, Bill Yuchen  and
      Park, Chan Young  and
      Li, Shuyue Stella  and
      Ravi, Sahithya  and
      Bhatia, Mehar  and
      Antoniak, Maria  and
      Tsvetkov, Yulia  and
      Shwartz, Vered  and
      Choi, Yejin",
    editor = "Che, Wanxiang  and
      Nabende, Joyce  and
      Shutova, Ekaterina  and
      Pilehvar, Mohammad Taher",
    booktitle = "Proceedings of the 63rd Annual Meeting of the Association for Computational Linguistics (Volume 1: Long Papers)",
    month = jul,
    year = "2025",
    address = "Vienna, Austria",
    publisher = "Association for Computational Linguistics",
    url = "https://aclanthology.org/2025.acl-long.1247/",
    doi = "10.18653/v1/2025.acl-long.1247",
    pages = "25663--25701",
    ISBN = "979-8-89176-251-0"
}

@inproceedings{robinson-etal-2025-al,
    title = "{AL}-{QASIDA}: Analyzing {LLM} Quality and Accuracy Systematically in Dialectal {A}rabic",
    author = "Robinson, Nathaniel Romney  and
      Abdelmoneim, Shahd  and
      Marchisio, Kelly  and
      Ruder, Sebastian",
    editor = "Che, Wanxiang  and
      Nabende, Joyce  and
      Shutova, Ekaterina  and
      Pilehvar, Mohammad Taher",
    booktitle = "Findings of the Association for Computational Linguistics: ACL 2025",
    month = jul,
    year = "2025",
    address = "Vienna, Austria",
    publisher = "Association for Computational Linguistics",
    url = "https://aclanthology.org/2025.findings-acl.1137/",
    doi = "10.18653/v1/2025.findings-acl.1137",
    pages = "22048--22065",
    ISBN = "979-8-89176-256-5"
}

@misc{ma2025hungergamedebateemergence,
      title={The Hunger Game Debate: On the Emergence of Over-Competition in Multi-Agent Systems}, 
      author={Xinbei Ma and Ruotian Ma and Xingyu Chen and Zhengliang Shi and Mengru Wang and Jen-tse Huang and Qu Yang and Wenxuan Wang and Fanghua Ye and Qingxuan Jiang and Mengfei Zhou and Zhuosheng Zhang and Rui Wang and Hai Zhao and Zhaopeng Tu and Xiaolong Li and Linus},
      year={2025},
      eprint={2509.26126},
      archivePrefix={arXiv},
      primaryClass={cs.CL},
      url={https://arxiv.org/abs/2509.26126}, 
}

@misc{comanici2025gemini25pushingfrontier,
      title={Gemini 2.5: Pushing the Frontier with Advanced Reasoning, Multimodality, Long Context, and Next Generation Agentic Capabilities}, 
      author={Gheorghe Comanici and Eric Bieber and Mike Schaekermann and Ice Pasupat and Noveen Sachdeva and Inderjit Dhillon and Marcel Blistein and Ori Ram and Dan Zhang and Evan Rosen and Luke Marris and Sam Petulla and Colin Gaffney and Asaf Aharoni and Nathan Lintz and Tiago Cardal Pais and Henrik Jacobsson and Idan Szpektor and Nan-Jiang Jiang and Krishna Haridasan and Ahmed Omran and Nikunj Saunshi and Dara Bahri and Gaurav Mishra and Eric Chu and Toby Boyd and Brad Hekman and Aaron Parisi and Chaoyi Zhang and Kornraphop Kawintiranon and Tania Bedrax-Weiss and Oliver Wang and Ya Xu and Ollie Purkiss and Uri Mendlovic and Ilaï Deutel and Nam Nguyen and Adam Langley and Flip Korn and Lucia Rossazza and Alexandre Ramé and Sagar Waghmare and Helen Miller and Nathan Byrd and Ashrith Sheshan and Raia Hadsell and Sangnie Bhardwaj and Pawel Janus and Tero Rissa and Dan Horgan and Alvin Abdagic and Lior Belenki and James Allingham and Anima Singh and Theo Guidroz and Srivatsan Srinivasan and Herman Schmit and Kristen Chiafullo and Andre Elisseeff and Nilpa Jha and Prateek Kolhar and Leonard Berrada and Frank Ding and Xiance Si and Shrestha Basu Mallick and Franz Och and Sofia Erell and Eric Ni and Tejasi Latkar and Sherry Yang and Petar Sirkovic and Ziqiang Feng and Robert Leland and Rachel Hornung and Gang Wu and Charles Blundell and Hamidreza Alvari and Po-Sen Huang and Cathy Yip and Sanja Deur and Li Liu and Gabriela Surita and Pablo Duque and Dima Damen and Johnson Jia and Arthur Guez and Markus Mircea and Animesh Sinha and Alberto Magni and Paweł Stradomski and Tal Marian and Vlado Galić and Wenhu Chen and Hisham Husain and Achintya Singhal and Dominik Grewe and François-Xavier Aubet and Shuang Song and Lorenzo Blanco and Leland Rechis and Lewis Ho and Rich Munoz and Kelvin Zheng and Jessica Hamrick and Kevin Mather and Hagai Taitelbaum and Eliza Rutherford and Yun Lei and Kuangyuan Chen and Anand Shukla and Erica Moreira and Eric Doi and Berivan Isik and Nir Shabat and Dominika Rogozińska and Kashyap Kolipaka and Jason Chang and Eugen Vušak and Srinivasan Venkatachary and Shadi Noghabi and Tarun Bharti and Younghoon Jun and Aleksandr Zaks and Simon Green and Jeshwanth Challagundla and William Wong and Muqthar Mohammad and Dean Hirsch and Yong Cheng and Iftekhar Naim and Lev Proleev and Damien Vincent and Aayush Singh and Maxim Krikun and Dilip Krishnan and Zoubin Ghahramani and Aviel Atias and Rajeev Aggarwal and Christo Kirov and Dimitrios Vytiniotis and Christy Koh and Alexandra Chronopoulou and Pawan Dogra and Vlad-Doru Ion and Gladys Tyen and Jason Lee and Felix Weissenberger and Trevor Strohman and Ashwin Balakrishna and Jack Rae and Marko Velic and Raoul de Liedekerke and Oded Elyada and Wentao Yuan and Canoee Liu and Lior Shani and Sergey Kishchenko and Bea Alessio and Yandong Li and Richard Song and Sam Kwei and Orion Jankowski and Aneesh Pappu and Youhei Namiki and Yenai Ma and Nilesh Tripuraneni and Colin Cherry and Marissa Ikonomidis and Yu-Cheng Ling and Colin Ji and Beka Westberg and Auriel Wright and Da Yu and David Parkinson and Swaroop Ramaswamy and Jerome Connor and Soheil Hassas Yeganeh and Snchit Grover and George Kenwright and Lubo Litchev and Chris Apps and Alex Tomala and Felix Halim and Alex Castro-Ros and Zefei Li and Anudhyan Boral and Pauline Sho and Michal Yarom and Eric Malmi and David Klinghoffer and Rebecca Lin and Alan Ansell and Pradeep Kumar S and Shubin Zhao and Siqi Zuo and Adam Santoro and Heng-Tze Cheng and Solomon Demmessie and Yuchi Liu and Nicole Brichtova and Allie Culp and Nathaniel Braun and Dan Graur and Will Ng and Nikhil Mehta and Aaron Phillips and Patrik Sundberg and Varun Godbole and Fangyu Liu and Yash Katariya and David Rim and Mojtaba Seyedhosseini and Sean Ammirati and Jonas Valfridsson and Mahan Malihi and Timothy Knight and Andeep Toor and Thomas Lampe and Abe Ittycheriah and Lewis Chiang and Chak Yeung and Alexandre Fréchette and Jinmeng Rao and Huisheng Wang and Himanshu Srivastava and Richard Zhang and Rocky Rhodes and Ariel Brand and Dean Weesner and Ilya Figotin and Felix Gimeno and Rachana Fellinger and Pierre Marcenac and José Leal and Eyal Marcus and Victor Cotruta and Rodrigo Cabrera and Sheryl Luo and Dan Garrette and Vera Axelrod and Sorin Baltateanu and David Barker and Dongkai Chen and Horia Toma and Ben Ingram and Jason Riesa and Chinmay Kulkarni and Yujing Zhang and Hongbin Liu and Chao Wang and Martin Polacek and Will Wu and Kai Hui and Adrian N Reyes and Yi Su and Megan Barnes and Ishaan Malhi and Anfal Siddiqui and Qixuan Feng and Mihai Damaschin and Daniele Pighin and Andreas Steiner and Samuel Yang and Ramya Sree Boppana and Simeon Ivanov and Arun Kandoor and Aditya Shah and Asier Mujika and Da Huang and Christopher A. Choquette-Choo and Mohak Patel and Tianhe Yu and Toni Creswell and Jerry and Liu and Catarina Barros and Yasaman Razeghi and Aurko Roy and Phil Culliton and Binbin Xiong and Jiaqi Pan and Thomas Strohmann and Tolly Powell and Babi Seal and Doug DeCarlo and Pranav Shyam and Kaan Katircioglu and Xuezhi Wang and Cassidy Hardin and Immanuel Odisho and Josef Broder and Oscar Chang and Arun Nair and Artem Shtefan and Maura O'Brien and Manu Agarwal and Sahitya Potluri and Siddharth Goyal and Amit Jhindal and Saksham Thakur and Yury Stuken and James Lyon and Kristina Toutanova and Fangxiaoyu Feng and Austin Wu and Ben Horn and Alek Wang and Alex Cullum and Gabe Taubman and Disha Shrivastava and Chongyang Shi and Hamish Tomlinson and Roma Patel and Tao Tu and Ada Maksutaj Oflazer and Francesco Pongetti and Mingyao Yang and Adrien Ali Taïga and Vincent Perot and Nuo Wang Pierse and Feng Han and Yoel Drori and Iñaki Iturrate and Ayan Chakrabarti and Legg Yeung and Dave Dopson and Yi-ting Chen and Apoorv Kulshreshtha and Tongfei Guo and Philip Pham and Tal Schuster and Junquan Chen and Alex Polozov and Jinwei Xing and Huanjie Zhou and Praneeth Kacham and Doron Kukliansky and Antoine Miech and Sergey Yaroshenko and Ed Chi and Sholto Douglas and Hongliang Fei and Mathieu Blondel and Preethi Myla and Lior Madmoni and Xing Wu and Daniel Keysers and Kristian Kjems and Isabela Albuquerque and Lijun Yu and Joel D'sa and Michelle Plantan and Vlad Ionescu and Jaume Sanchez Elias and Abhirut Gupta and Manish Reddy Vuyyuru and Fred Alcober and Tong Zhou and Kaiyang Ji and Florian Hartmann and Subha Puttagunta and Hugo Song and Ehsan Amid and Anca Stefanoiu and Andrew Lee and Paul Pucciarelli and Emma Wang and Amit Raul and Slav Petrov and Isaac Tian and Valentin Anklin and Nana Nti and Victor Gomes and Max Schumacher and Grace Vesom and Alex Panagopoulos and Konstantinos Bousmalis and Daniel Andor and Josh Jacob and Yuan Zhang and Bill Rosgen and Matija Kecman and Matthew Tung and Alexandra Belias and Noah Goodman and Paul Covington and Brian Wieder and Nikita Saxena and Elnaz Davoodi and Muhuan Huang and Sharath Maddineni and Vincent Roulet and Folawiyo Campbell-Ajala and Pier Giuseppe Sessa and Xintian and Wu and Guangda Lai and Paul Collins and Alex Haig and Vytenis Sakenas and Xiaowei Xu and Marissa Giustina and Laurent El Shafey and Pichi Charoenpanit and Shefali Garg and Joshua Ainslie and Boone Severson and Montse Gonzalez Arenas and Shreya Pathak and Sujee Rajayogam and Jie Feng and Michiel Bakker and Sheng Li and Nevan Wichers and Jamie Rogers and Xinyang Geng and Yeqing Li and Rolf Jagerman and Chao Jia and Nadav Olmert and David Sharon and Matthew Mauger and Sandeep Mariserla and Hongxu Ma and Megha Mohabey and Kyuyeun Kim and Alek Andreev and Scott Pollom and Juliette Love and Vihan Jain and Priyanka Agrawal and Yannick Schroecker and Alisa Fortin and Manfred Warmuth and Ji Liu and Andrew Leach and Irina Blok and Ganesh Poomal Girirajan and Roee Aharoni and Benigno Uria and Andrei Sozanschi and Dan Goldberg and Lucian Ionita and Marco Tulio Ribeiro and Martin Zlocha and Vighnesh Birodkar and Sami Lachgar and Liangzhe Yuan and Himadri Choudhury and Matt Ginsberg and Fei Zheng and Gregory Dibb and Emily Graves and Swachhand Lokhande and Gabriel Rasskin and George-Cristian Muraru and Corbin Quick and Sandeep Tata and Pierre Sermanet and Aditya Chawla and Itay Karo and Yan Wang and Susan Zhang and Orgad Keller and Anca Dragan and Guolong Su and Ian Chou and Xi Liu and Yiqing Tao and Shruthi Prabhakara and Marc Wilson and Ruibo Liu and Shibo Wang and Georgie Evans and David Du and Alfonso Castaño and Gautam Prasad and Mona El Mahdy and Sebastian Gerlach and Machel Reid and Jarrod Kahn and Amir Zait and Thanumalayan Sankaranarayana Pillai and Thatcher Ulrich and Guanyu Wang and Jan Wassenberg and Efrat Farkash and Kiran Yalasangi and Congchao Wang and Maria Bauza and Simon Bucher and Ting Liu and Jun Yan and Gary Leung and Vikas Sindhwani and Parker Barnes and Avi Singh and Ivan Jurin and Jichuan Chang and Niket Kumar Bhumihar and Sivan Eiger and Gui Citovsky and Ben Withbroe and Zhang Li and Siyang Xue and Niccolò Dal Santo and Georgi Stoyanov and Yves Raimond and Steven Zheng and Yilin Gao and Vít Listík and Sławek Kwasiborski and Rachel Saputro and Adnan Ozturel and Ganesh Mallya and Kushal Majmundar and Ross West and Paul Caron and Jinliang Wei and Lluis Castrejon and Sharad Vikram and Deepak Ramachandran and Nikhil Dhawan and Jiho Park and Sara Smoot and George van den Driessche and Yochai Blau and Chase Malik and Wei Liang and Roy Hirsch and Cicero Nogueira dos Santos and Eugene Weinstein and Aäron van den Oord and Sid Lall and Nicholas FitzGerald and Zixuan Jiang and Xuan Yang and Dale Webster and Ali Elqursh and Aedan Pope and Georges Rotival and David Raposo and Wanzheng Zhu and Jeff Dean and Sami Alabed and Dustin Tran and Arushi Gupta and Zach Gleicher and Jessica Austin and Edouard Rosseel and Megh Umekar and Dipanjan Das and Yinghao Sun and Kai Chen and Karolis Misiunas and Xiang Zhou and Yixian Di and Alyssa Loo and Josh Newlan and Bo Li and Vinay Ramasesh and Ying Xu and Alex Chen and Sudeep Gandhe and Radu Soricut and Nikita Gupta and Shuguang Hu and Seliem El-Sayed and Xavier Garcia and Idan Brusilovsky and Pu-Chin Chen and Andrew Bolt and Lu Huang and Alex Gurney and Zhiying Zhang and Alexander Pritzel and Jarek Wilkiewicz and Bryan Seybold and Bhargav Kanagal Shamanna and Felix Fischer and Josef Dean and Karan Gill and Ross Mcilroy and Abhishek Bhowmick and Jeremy Selier and Antoine Yang and Derek Cheng and Vladimir Magay and Jie Tan and Dhriti Varma and Christian Walder and Tomas Kocisky and Ryo Nakashima and Paul Natsev and Mike Kwong and Ionel Gog and Chiyuan Zhang and Sander Dieleman and Thomas Jimma and Andrey Ryabtsev and Siddhartha Brahma and David Steiner and Dayou Du and Ante Žužul and Mislav Žanić and Mukund Raghavachari and Willi Gierke and Zeyu Zheng and Dessie Petrova and Yann Dauphin and Yuchuan Liu and Ido Kessler and Steven Hand and Chris Duvarney and Seokhwan Kim and Hyo Lee and Léonard Hussenot and Jeffrey Hui and Josh Smith and Deepali Jain and Jiawei Xia and Gaurav Singh Tomar and Keyvan Amiri and Du Phan and Fabian Fuchs and Tobias Weyand and Nenad Tomasev and Alexandra Cordell and Xin Liu and Jonathan Mallinson and Pankaj Joshi and Andy Crawford and Arun Suggala and Steve Chien and Nick Fernando and Mariella Sanchez-Vargas and Duncan Williams and Phil Crone and Xiyang Luo and Igor Karpov and Jyn Shan and Terry Thurk and Robin Strudel and Paul Voigtlaender and Piyush Patil and Tim Dozat and Ali Khodaei and Sahil Singla and Piotr Ambroszczyk and Qiyin Wu and Yifan Chang and Brian Roark and Chaitra Hegde and Tianli Ding and Angelos Filos and Zhongru Wu and André Susano Pinto and Shuang Liu and Saarthak Khanna and Aditya Pandey and Siobhan Mcloughlin and Qiujia Li and Sam Haves and Allan Zhou and Elena Buchatskaya and Isabel Leal and Peter de Boursac and Nami Akazawa and Nina Anderson and Terry Chen and Krishna Somandepalli and Chen Liang and Sheela Goenka and Stephanie Winkler and Alexander Grushetsky and Yifan Ding and Jamie Smith and Fan Ye and Jordi Pont-Tuset and Eric Li and Ruichao Li and Tomer Golany and Dawid Wegner and Tao Jiang and Omer Barak and Yuan Shangguan and Eszter Vértes and Renee Wong and Jörg Bornschein and Alex Tudor and Michele Bevilacqua and Tom Schaul and Ankit Singh Rawat and Yang Zhao and Kyriakos Axiotis and Lei Meng and Cory McLean and Jonathan Lai and Jennifer Beattie and Nate Kushman and Yaxin Liu and Blair Kutzman and Fiona Lang and Jingchen Ye and Praneeth Netrapalli and Pushkar Mishra and Myriam Khan and Megha Goel and Rob Willoughby and David Tian and Honglei Zhuang and JD Chen and Zak Tsai and Tasos Kementsietsidis and Arjun Khare and James Keeling and Keyang Xu and Nathan Waters and Florent Altché and Ashok Popat and Bhavishya Mittal and David Saxton and Dalia El Badawy and Michael Mathieu and Zheng Zheng and Hao Zhou and Nishant Ranka and Richard Shin and Qingnan Duan and Tim Salimans and Ioana Mihailescu and Uri Shaham and Ming-Wei Chang and Yannis Assael and Nishanth Dikkala and Martin Izzard and Vincent Cohen-Addad and Cat Graves and Vlad Feinberg and Grace Chung and DJ Strouse and Danny Karmon and Sahand Sharifzadeh and Zoe Ashwood and Khiem Pham and Jon Blanton and Alex Vasiloff and Jarred Barber and Mark Geller and Aurick Zhou and Fedir Zubach and Tzu-Kuo Huang and Lei Zhang and Himanshu Gupta and Matt Young and Julia Proskurnia and Ronny Votel and Valentin Gabeur and Gabriel Barcik and Aditya Tripathi and Hongkun Yu and Geng Yan and Beer Changpinyo and Filip Pavetić and Amy Coyle and Yasuhisa Fujii and Jorge Gonzalez Mendez and Tianhao Zhou and Harish Rajamani and Blake Hechtman and Eddie Cao and Da-Cheng Juan and Yi-Xuan Tan and Valentin Dalibard and Yilun Du and Natalie Clay and Kaisheng Yao and Wenhao Jia and Dimple Vijaykumar and Yuxiang Zhou and Xinyi Bai and Wei-Chih Hung and Steven Pecht and Georgi Todorov and Nikhil Khadke and Pramod Gupta and Preethi Lahoti and Arnaud Autef and Karthik Duddu and James Lee-Thorp and Alexander Bykovsky and Tautvydas Misiunas and Sebastian Flennerhag and Santhosh Thangaraj and Jed McGiffin and Zack Nado and Markus Kunesch and Andreas Noever and Amir Hertz and Marco Liang and Victor Stone and Evan Palmer and Samira Daruki and Arijit Pramanik and Siim Põder and Austin Kyker and Mina Khan and Evgeny Sluzhaev and Marvin Ritter and Avraham Ruderman and Wenlei Zhou and Chirag Nagpal and Kiran Vodrahalli and George Necula and Paul Barham and Ellie Pavlick and Jay Hartford and Izhak Shafran and Long Zhao and Maciej Mikuła and Tom Eccles and Hidetoshi Shimokawa and Kanav Garg and Luke Vilnis and Hanwen Chen and Ilia Shumailov and Kuang-Huei Lee and Abdelrahman Abdelhamed and Meiyan Xie and Vered Cohen and Ester Hlavnova and Dan Malkin and Chawin Sitawarin and James Lottes and Pauline Coquinot and Tianli Yu and Sandeep Kumar and Jingwei Zhang and Aroma Mahendru and Zafarali Ahmed and James Martens and Tao Chen and Aviel Boag and Daiyi Peng and Coline Devin and Arseniy Klimovskiy and Mary Phuong and Danny Vainstein and Jin Xie and Bhuvana Ramabhadran and Nathan Howard and Xinxin Yu and Gitartha Goswami and Jingyu Cui and Sam Shleifer and Mario Pinto and Chih-Kuan Yeh and Ming-Hsuan Yang and Sara Javanmardi and Dan Ethier and Chace Lee and Jordi Orbay and Suyog Kotecha and Carla Bromberg and Pete Shaw and James Thornton and Adi Gerzi Rosenthal and Shane Gu and Matt Thomas and Ian Gemp and Aditya Ayyar and Asahi Ushio and Aarush Selvan and Joel Wee and Chenxi Liu and Maryam Majzoubi and Weiren Yu and Jake Abernethy and Tyler Liechty and Renke Pan and Hoang Nguyen and Qiong and Hu and Sarah Perrin and Abhinav Arora and Emily Pitler and Weiyi Wang and Kaushik Shivakumar and Flavien Prost and Ben Limonchik and Jing Wang and Yi Gao and Timothee Cour and Shyamal Buch and Huan Gui and Maria Ivanova and Philipp Neubeck and Kelvin Chan and Lucy Kim and Huizhong Chen and Naman Goyal and Da-Woon Chung and Lu Liu and Yao Su and Anastasia Petrushkina and Jiajun Shen and Armand Joulin and Yuanzhong Xu and Stein Xudong Lin and Yana Kulizhskaya and Ciprian Chelba and Shobha Vasudevan and Eli Collins and Vasilisa Bashlovkina and Tony Lu and Doug Fritz and Jongbin Park and Yanqi Zhou and Chen Su and Richard Tanburn and Mikhail Sushkov and Mitchelle Rasquinha and Jinning Li and Jennifer Prendki and Yiming Li and Pallavi LV and Shriya Sharma and Hen Fitoussi and Hui Huang and Andrew Dai and Phuong Dao and Mike Burrows and Henry Prior and Danfeng Qin and Golan Pundak and Lars Lowe Sjoesund and Art Khurshudov and Zhenkai Zhu and Albert Webson and Elizabeth Kemp and Tat Tan and Saurabh Agrawal and Susie Sargsyan and Liqun Cheng and Jim Stephan and Tom Kwiatkowski and David Reid and Arunkumar Byravan and Assaf Hurwitz Michaely and Nicolas Heess and Luowei Zhou and Sonam Goenka and Viral Carpenter and Anselm Levskaya and Bo Wang and Reed Roberts and Rémi Leblond and Sharat Chikkerur and Stav Ginzburg and Max Chang and Robert Riachi and Chuqiao and Xu and Zalán Borsos and Michael Pliskin and Julia Pawar and Morgane Lustman and Hannah Kirkwood and Ankit Anand and Aditi Chaudhary and Norbert Kalb and Kieran Milan and Sean Augenstein and Anna Goldie and Laurel Prince and Karthik Raman and Yanhua Sun and Vivian Xia and Aaron Cohen and Zhouyuan Huo and Josh Camp and Seher Ellis and Lukas Zilka and David Vilar Torres and Lisa Patel and Sho Arora and Betty Chan and Jonas Adler and Kareem Ayoub and Jacky Liang and Fayaz Jamil and Jiepu Jiang and Simon Baumgartner and Haitian Sun and Yael Karov and Yaroslav Akulov and Hui Zheng and Irene Cai and Claudio Fantacci and James Rubin and Alex Rav Acha and Mengchao Wang and Nina D'Souza and Rohit Sathyanarayana and Shengyang Dai and Simon Rowe and Andrey Simanovsky and Omer Goldman and Yuheng Kuang and Xiaoyue Pan and Andrew Rosenberg and Tania Rojas-Esponda and Praneet Dutta and Amy Zeng and Irina Jurenka and Greg Farquhar and Yamini Bansal and Shariq Iqbal and Becca Roelofs and Ga-Young Joung and Parker Beak and Changwan Ryu and Ryan Poplin and Yan Wu and Jean-Baptiste Alayrac and Senaka Buthpitiya and Olaf Ronneberger and Caleb Habtegebriel and Wei Li and Paul Cavallaro and Aurora Wei and Guy Bensky and Timo Denk and Harish Ganapathy and Jeff Stanway and Pratik Joshi and Francesco Bertolini and Jessica Lo and Olivia Ma and Zachary Charles and Geta Sampemane and Himanshu Sahni and Xu Chen and Harry Askham and David Gaddy and Peter Young and Jiewen Tan and Matan Eyal and Arthur Bražinskas and Li Zhong and Zhichun Wu and Mark Epstein and Kai Bailey and Andrew Hard and Kamyu Lee and Sasha Goldshtein and Alex Ruiz and Mohammed Badawi and Matthias Lochbrunner and JK Kearns and Ashley Brown and Fabio Pardo and Theophane Weber and Haichuan Yang and Pan-Pan Jiang and Berkin Akin and Zhao Fu and Marcus Wainwright and Chi Zou and Meenu Gaba and Pierre-Antoine Manzagol and Wendy Kan and Yang Song and Karina Zainullina and Rui Lin and Jeongwoo Ko and Salil Deshmukh and Apoorv Jindal and James Svensson and Divya Tyam and Heri Zhao and Christine Kaeser-Chen and Scott Baird and Pooya Moradi and Jamie Hall and Qiuchen Guo and Vincent Tsang and Bowen Liang and Fernando Pereira and Suhas Ganesh and Ivan Korotkov and Jakub Adamek and Sridhar Thiagarajan and Vinh Tran and Charles Chen and Chris Tar and Sanil Jain and Ishita Dasgupta and Taylan Bilal and David Reitter and Kai Zhao and Giulia Vezzani and Yasmin Gehman and Pulkit Mehta and Lauren Beltrone and Xerxes Dotiwalla and Sergio Guadarrama and Zaheer Abbas and Stefani Karp and Petko Georgiev and Chun-Sung Ferng and Marc Brockschmidt and Liqian Peng and Christoph Hirnschall and Vikas Verma and Yingying Bi and Ying Xiao and Avigail Dabush and Kelvin Xu and Phil Wallis and Randall Parker and Qifei Wang and Yang Xu and Ilkin Safarli and Dinesh Tewari and Yin Zhang and Seungyeon Kim and Andrea Gesmundo and Mackenzie Thomas and Sergey Levi and Ahmed Chowdhury and Kanishka Rao and Peter Garst and Sam Conway-Rahman and Helen Ran and Kay McKinney and Zhisheng Xiao and Wenhao Yu and Rohan Agrawal and Axel Stjerngren and Catalin Ionescu and Jingjing Chen and Vivek Sharma and Justin Chiu and Fei Liu and Ken Franko and Clayton Sanford and Xingyu Cai and Paul Michel and Sanjay Ganapathy and Jane Labanowski and Zachary Garrett and Ben Vargas and Sean Sun and Bryan Gale and Thomas Buschmann and Guillaume Desjardins and Nimesh Ghelani and Palak Jain and Mudit Verma and Chulayuth Asawaroengchai and Julian Eisenschlos and Jitendra Harlalka and Hideto Kazawa and Don Metzler and Joshua Howland and Ying Jian and Jake Ades and Viral Shah and Tynan Gangwani and Seungji Lee and Roman Ring and Steven M. Hernandez and Dean Reich and Amer Sinha and Ashutosh Sathe and Joe Kovac and Ashleah Gill and Ajay Kannan and Andrea D'olimpio and Martin Sevenich and Jay Whang and Been Kim and Khe Chai Sim and Jilin Chen and Jiageng Zhang and Shuba Lall and Yossi Matias and Bill Jia and Abe Friesen and Sara Nasso and Ashish Thapliyal and Bryan Perozzi and Ting Yu and Anna Shekhawat and Safeen Huda and Peter Grabowski and Eric Wang and Ashwin Sreevatsa and Hilal Dib and Mehadi Hassen and Parker Schuh and Vedrana Milutinovic and Chris Welty and Michael Quinn and Ali Shah and Bangju Wang and Gabe Barth-Maron and Justin Frye and Natalie Axelsson and Tao Zhu and Yukun Ma and Irene Giannoumis and Hanie Sedghi and Chang Ye and Yi Luan and Kevin Aydin and Bilva Chandra and Vivek Sampathkumar and Ronny Huang and Victor Lavrenko and Ahmed Eleryan and Zhi Hong and Steven Hansen and Sara Mc Carthy and Bidisha Samanta and Domagoj Ćevid and Xin Wang and Fangtao Li and Michael Voznesensky and Matt Hoffman and Andreas Terzis and Vikash Sehwag and Gil Fidel and Luheng He and Mu Cai and Yanzhang He and Alex Feng and Martin Nikoltchev and Samrat Phatale and Jason Chase and Rory Lawton and Ming Zhang and Tom Ouyang and Manuel Tragut and Mehdi Hafezi Manshadi and Arjun Narayanan and Jiaming Shen and Xu Gao and Tolga Bolukbasi and Nick Roy and Xin Li and Daniel Golovin and Liviu Panait and Zhen Qin and Guangxing Han and Thomas Anthony and Sneha Kudugunta and Viorica Patraucean and Aniket Ray and Xinyun Chen and Xiaochen Yang and Tanuj Bhatia and Pranav Talluri and Alex Morris and Andrija Ražnatović and Bethanie Brownfield and James An and Sheng Peng and Patrick Kane and Ce Zheng and Nico Duduta and Joshua Kessinger and James Noraky and Siqi Liu and Keran Rong and Petar Veličković and Keith Rush and Alex Goldin and Fanny Wei and Shiva Mohan Reddy Garlapati and Caroline Pantofaru and Okwan Kwon and Jianmo Ni and Eric Noland and Julia Di Trapani and Françoise Beaufays and Abhijit Guha Roy and Yinlam Chow and Aybuke Turker and Geoffrey Cideron and Lantao Mei and Jon Clark and Qingyun Dou and Matko Bošnjak and Ralph Leith and Yuqing Du and Amir Yazdanbakhsh and Milad Nasr and Chester Kwak and Suraj Satishkumar Sheth and Alex Kaskasoli and Ankesh Anand and Balaji Lakshminarayanan and Sammy Jerome and David Bieber and Chun-Te Chu and Alexandre Senges and Tianxiao Shen and Mukund Sridhar and Ndaba Ndebele and Benjamin Beyret and Shakir Mohamed and Mia Chen and Markus Freitag and Jiaxian Guo and Luyang Liu and Paul Roit and Heng Chen and Shen Yan and Tom Stone and JD Co-Reyes and Jeremy Cole and Salvatore Scellato and Shekoofeh Azizi and Hadi Hashemi and Alicia Jin and Anand Iyer and Marcella Valentine and András György and Arun Ahuja and Daniel Hernandez Diaz and Chen-Yu Lee and Nathan Clement and Weize Kong and Drew Garmon and Ishaan Watts and Kush Bhatia and Khyatti Gupta and Matt Miecnikowski and Hugo Vallet and Ankur Taly and Edward Loper and Saket Joshi and James Atwood and Jo Chick and Mark Collier and Fotis Iliopoulos and Ryan Trostle and Beliz Gunel and Ramiro Leal-Cavazos and Arnar Mar Hrafnkelsson and Michael Guzman and Xiaoen Ju and Andy Forbes and Jesse Emond and Kushal Chauhan and Ben Caine and Li Xiao and Wenjun Zeng and Alexandre Moufarek and Daniel Murphy and Maya Meng and Nitish Gupta and Felix Riedel and Anil Das and Elijah Lawal and Shashi Narayan and Tiberiu Sosea and James Swirhun and Linda Friso and Behnam Neyshabur and Jing Lu and Sertan Girgin and Michael Wunder and Edouard Yvinec and Aroonalok Pyne and Victor Carbune and Shruti Rijhwani and Yang Guo and Tulsee Doshi and Anton Briukhov and Max Bain and Ayal Hitron and Xuanhui Wang and Ashish Gupta and Ke Chen and Cosmo Du and Weiyang Zhang and Dhruv Shah and Arjun Akula and Max Dylla and Ashyana Kachra and Weicheng Kuo and Tingting Zou and Lily Wang and Luyao Xu and Jifan Zhu and Justin Snyder and Sachit Menon and Orhan Firat and Igor Mordatch and Yuan Yuan and Natalia Ponomareva and Rory Blevins and Lawrence Moore and Weijun Wang and Phil Chen and Martin Scholz and Artur Dwornik and Jason Lin and Sicheng Li and Diego Antognini and Te I and Xiaodan Song and Matt Miller and Uday Kalra and Adam Raveret and Oscar Akerlund and Felix Wu and Andrew Nystrom and Namrata Godbole and Tianqi Liu and Hannah DeBalsi and Jewel Zhao and Buhuang Liu and Avi Caciularu and Lauren Lax and Urvashi Khandelwal and Victoria Langston and Eric Bailey and Silvio Lattanzi and Yufei Wang and Neel Kovelamudi and Sneha Mondal and Guru Guruganesh and Nan Hua and Ofir Roval and Paweł Wesołowski and Rishikesh Ingale and Jonathan Halcrow and Tim Sohn and Christof Angermueller and Bahram Raad and Eli Stickgold and Eva Lu and Alec Kosik and Jing Xie and Timothy Lillicrap and Austin Huang and Lydia Lihui Zhang and Dominik Paulus and Clement Farabet and Alex Wertheim and Bing Wang and Rishabh Joshi and Chu-ling Ko and Yonghui Wu and Shubham Agrawal and Lily Lin and XiangHai Sheng and Peter Sung and Tyler Breland-King and Christina Butterfield and Swapnil Gawde and Sumeet Singh and Qiao Zhang and Raj Apte and Shilpa Shetty and Adrian Hutter and Tao Li and Elizabeth Salesky and Federico Lebron and Jonni Kanerva and Michela Paganini and Arthur Nguyen and Rohith Vallu and Jan-Thorsten Peter and Sarmishta Velury and David Kao and Jay Hoover and Anna Bortsova and Colton Bishop and Shoshana Jakobovits and Alessandro Agostini and Alekh Agarwal and Chang Liu and Charles Kwong and Sasan Tavakkol and Ioana Bica and Alex Greve and Anirudh GP and Jake Marcus and Le Hou and Tom Duerig and Rivka Moroshko and Dave Lacey and Andy Davis and Julien Amelot and Guohui Wang and Frank Kim and Theofilos Strinopoulos and Hui Wan and Charline Le Lan and Shankar Krishnan and Haotian Tang and Peter Humphreys and Junwen Bai and Idan Heimlich Shtacher and Diego Machado and Chenxi Pang and Ken Burke and Dangyi Liu and Renga Aravamudhan and Yue Song and Ed Hirst and Abhimanyu Singh and Brendan Jou and Liang Bai and Francesco Piccinno and Chuyuan Kelly Fu and Robin Alazard and Barak Meiri and Daniel Winter and Charlie Chen and Mingda Zhang and Jens Heitkaemper and John Lambert and Jinhyuk Lee and Alexander Frömmgen and Sergey Rogulenko and Pranav Nair and Paul Niemczyk and Anton Bulyenov and Bibo Xu and Hadar Shemtov and Morteza Zadimoghaddam and Serge Toropov and Mateo Wirth and Hanjun Dai and Sreenivas Gollapudi and Daniel Zheng and Alex Kurakin and Chansoo Lee and Kalesha Bullard and Nicolas Serrano and Ivana Balazevic and Yang Li and Johan Schalkwyk and Mark Murphy and Mingyang Zhang and Kevin Sequeira and Romina Datta and Nishant Agrawal and Charles Sutton and Nithya Attaluri and Mencher Chiang and Wael Farhan and Gregory Thornton and Kate Lin and Travis Choma and Hung Nguyen and Kingshuk Dasgupta and Dirk Robinson and Iulia Comşa and Michael Riley and Arjun Pillai and Basil Mustafa and Ben Golan and Amir Zandieh and Jean-Baptiste Lespiau and Billy Porter and David Ross and Sujeevan Rajayogam and Mohit Agarwal and Subhashini Venugopalan and Bobak Shahriari and Qiqi Yan and Hao Xu and Taylor Tobin and Pavel Dubov and Hongzhi Shi and Adrià Recasens and Anton Kovsharov and Sebastian Borgeaud and Lucio Dery and Shanthal Vasanth and Elena Gribovskaya and Linhai Qiu and Mahdis Mahdieh and Wojtek Skut and Elizabeth Nielsen and CJ Zheng and Adams Yu and Carrie Grimes Bostock and Shaleen Gupta and Aaron Archer and Chris Rawles and Elinor Davies and Alexey Svyatkovskiy and Tomy Tsai and Yoni Halpern and Christian Reisswig and Bartek Wydrowski and Bo Chang and Joan Puigcerver and Mor Hazan Taege and Jian Li and Eva Schnider and Xinjian Li and Dragos Dena and Yunhan Xu and Umesh Telang and Tianze Shi and Heiga Zen and Kyle Kastner and Yeongil Ko and Neesha Subramaniam and Aviral Kumar and Pete Blois and Zhuyun Dai and John Wieting and Yifeng Lu and Yoel Zeldes and Tian Xie and Anja Hauth and Alexandru Ţifrea and Yuqi Li and Sam El-Husseini and Dan Abolafia and Howard Zhou and Wen Ding and Sahra Ghalebikesabi and Carlos Guía and Andrii Maksai and Ágoston Weisz and Sercan Arik and Nick Sukhanov and Aga Świetlik and Xuhui Jia and Luo Yu and Weiyue Wang and Mark Brand and Dawn Bloxwich and Sean Kirmani and Zhe Chen and Alec Go and Pablo Sprechmann and Nithish Kannen and Alen Carin and Paramjit Sandhu and Isabel Edkins and Leslie Nooteboom and Jai Gupta and Loren Maggiore and Javad Azizi and Yael Pritch and Pengcheng Yin and Mansi Gupta and Danny Tarlow and Duncan Smith and Desi Ivanov and Mohammad Babaeizadeh and Ankita Goel and Satish Kambala and Grace Chu and Matej Kastelic and Michelle Liu and Hagen Soltau and Austin Stone and Shivani Agrawal and Min Kim and Kedar Soparkar and Srinivas Tadepalli and Oskar Bunyan and Rachel Soh and Arvind Kannan and DY Kim and Blake JianHang Chen and Afief Halumi and Sudeshna Roy and Yulong Wang and Olcan Sercinoglu and Gena Gibson and Sijal Bhatnagar and Motoki Sano and Daniel von Dincklage and Qingchun Ren and Blagoj Mitrevski and Mirek Olšák and Jennifer She and Carl Doersch and Jilei and Wang and Bingyuan Liu and Qijun Tan and Tamar Yakar and Tris Warkentin and Alex Ramirez and Carl Lebsack and Josh Dillon and Rajiv Mathews and Tom Cobley and Zelin Wu and Zhuoyuan Chen and Jon Simon and Swaroop Nath and Tara Sainath and Alexei Bendebury and Ryan Julian and Bharath Mankalale and Daria Ćurko and Paulo Zacchello and Adam R. Brown and Kiranbir Sodhia and Heidi Howard and Sergi Caelles and Abhinav Gupta and Gareth Evans and Anna Bulanova and Lesley Katzen and Roman Goldenberg and Anton Tsitsulin and Joe Stanton and Benoit Schillings and Vitaly Kovalev and Corey Fry and Rushin Shah and Kuo Lin and Shyam Upadhyay and Cheng Li and Soroush Radpour and Marcello Maggioni and Jing Xiong and Lukas Haas and Jenny Brennan and Aishwarya Kamath and Nikolay Savinov and Arsha Nagrani and Trevor Yacovone and Ryan Kappedal and Kostas Andriopoulos and Li Lao and YaGuang Li and Grigory Rozhdestvenskiy and Kazuma Hashimoto and Andrew Audibert and Sophia Austin and Daniel Rodriguez and Anian Ruoss and Garrett Honke and Deep Karkhanis and Xi Xiong and Qing Wei and James Huang and Zhaoqi Leng and Vittal Premachandran and Stan Bileschi and Georgios Evangelopoulos and Thomas Mensink and Jay Pavagadhi and Denis Teplyashin and Paul Chang and Linting Xue and Garrett Tanzer and Sally Goldman and Kaushal Patel and Shixin Li and Jeremy Wiesner and Ivy Zheng and Ian Stewart-Binks and Jie Han and Zhi Li and Liangchen Luo and Karel Lenc and Mario Lučić and Fuzhao Xue and Ryan Mullins and Alexey Guseynov and Chung-Ching Chang and Isaac Galatzer-Levy and Adam Zhang and Garrett Bingham and Grace Hu and Ale Hartman and Yue Ma and Jordan Griffith and Alex Irpan and Carey Radebaugh and Summer Yue and Lijie Fan and Victor Ungureanu and Christina Sorokin and Hannah Teufel and Peiran Li and Rohan Anil and Dimitris Paparas and Todd Wang and Chu-Cheng Lin and Hui Peng and Megan Shum and Goran Petrovic and Demetra Brady and Richard Nguyen and Klaus Macherey and Zhihao Li and Harman Singh and Madhavi Yenugula and Mariko Iinuma and Xinyi Chen and Kavya Kopparapu and Alexey Stern and Shachi Dave and Chandu Thekkath and Florence Perot and Anurag Kumar and Fangda Li and Yang Xiao and Matthew Bilotti and Mohammad Hossein Bateni and Isaac Noble and Lisa Lee and Amelio Vázquez-Reina and Julian Salazar and Xiaomeng Yang and Boyu Wang and Ela Gruzewska and Anand Rao and Sindhu Raghuram and Zheng Xu and Eyal Ben-David and Jieru Mei and Sid Dalmia and Zhaoyi Zhang and Yuchen Liu and Gagan Bansal and Helena Pankov and Steven Schwarcz and Andrea Burns and Christine Chan and Sumit Sanghai and Ricky Liang and Ethan Liang and Antoine He and Amy Stuart and Arun Narayanan and Yukun Zhu and Christian Frank and Bahar Fatemi and Amit Sabne and Oran Lang and Indro Bhattacharya and Shane Settle and Maria Wang and Brendan McMahan and Andrea Tacchetti and Livio Baldini Soares and Majid Hadian and Serkan Cabi and Timothy Chung and Nikita Putikhin and Gang Li and Jeremy Chen and Austin Tarango and Henryk Michalewski and Mehran Kazemi and Hussain Masoom and Hila Sheftel and Rakesh Shivanna and Archita Vadali and Ramona Comanescu and Doug Reid and Joss Moore and Arvind Neelakantan and Michaël Sander and Jonathan Herzig and Aviv Rosenberg and Mostafa Dehghani and JD Choi and Michael Fink and Reid Hayes and Eric Ge and Shitao Weng and Chia-Hua Ho and John Karro and Kalpesh Krishna and Lam Nguyen Thiet and Amy Skerry-Ryan and Daniel Eppens and Marco Andreetto and Navin Sarma and Silvano Bonacina and Burcu Karagol Ayan and Megha Nawhal and Zhihao Shan and Mike Dusenberry and Shantanu Thakoor and Sagar Gubbi and Duc Dung Nguyen and Reut Tsarfaty and Samuel Albanie and Jovana Mitrović and Meet Gandhi and Bo-Juen Chen and Alessandro Epasto and Georgi Stephanov and Ye Jin and Samuel Gehman and Aida Amini and Jack Weber and Feryal Behbahani and Shawn Xu and Miltos Allamanis and Xi Chen and Myle Ott and Claire Sha and Michal Jastrzebski and Hang Qi and David Greene and Xinyi Wu and Abodunrinwa Toki and Daniel Vlasic and Jane Shapiro and Ragha Kotikalapudi and Zhe Shen and Takaaki Saeki and Sirui Xie and Albin Cassirer and Shikhar Bharadwaj and Tatsuya Kiyono and Srinadh Bhojanapalli and Elan Rosenfeld and Sam Ritter and Jieming Mao and João Gabriel Oliveira and Zoltan Egyed and Bernd Bandemer and Emilio Parisotto and Keisuke Kinoshita and Juliette Pluto and Petros Maniatis and Steve Li and Yaohui Guo and Golnaz Ghiasi and Jean Tarbouriech and Srimon Chatterjee and Julie Jin and Katrina and Xu and Jennimaria Palomaki and Séb Arnold and Madhavi Sewak and Federico Piccinini and Mohit Sharma and Ben Albrecht and Sean Purser-haskell and Ashwin Vaswani and Chongyan Chen and Matheus Wisniewski and Qin Cao and John Aslanides and Nguyet Minh Phu and Maximilian Sieb and Lauren Agubuzu and Anne Zheng and Daniel Sohn and Marco Selvi and Anders Andreassen and Krishan Subudhi and Prem Eruvbetine and Oliver Woodman and Tomas Mery and Sebastian Krause and Xiaoqi Ren and Xiao Ma and Jincheng Luo and Dawn Chen and Wei Fan and Henry Griffiths and Christian Schuler and Alice Li and Shujian Zhang and Jean-Michel Sarr and Shixin Luo and Riccardo Patana and Matthew Watson and Dani Naboulsi and Michael Collins and Sailesh Sidhwani and Emiel Hoogeboom and Sharon Silver and Emily Caveness and Xiaokai Zhao and Mikel Rodriguez and Maxine Deines and Libin Bai and Patrick Griffin and Marco Tagliasacchi and Emily Xue and Spandana Raj Babbula and Bo Pang and Nan Ding and Gloria Shen and Elijah Peake and Remi Crocker and Shubha Srinivas Raghvendra and Danny Swisher and Woohyun Han and Richa Singh and Ling Wu and Vladimir Pchelin and Tsendsuren Munkhdalai and Dana Alon and Geoff Bacon and Efren Robles and Jannis Bulian and Melvin Johnson and George Powell and Felipe Tiengo Ferreira and Yaoyiran Li and Frederik Benzing and Mihajlo Velimirović and Hubert Soyer and William Kong and Tony and Nguyên and Zhen Yang and Jeremiah Liu and Joost van Amersfoort and Daniel Gillick and Baochen Sun and Nathalie Rauschmayr and Katie Zhang and Serena Zhan and Tao Zhou and Alexey Frolov and Chengrun Yang and Denis Vnukov and Louis Rouillard and Hongji Li and Amol Mandhane and Nova Fallen and Rajesh Venkataraman and Clara Huiyi Hu and Jennifer Brennan and Jenny Lee and Jerry Chang and Martin Sundermeyer and Zhufeng Pan and Rosemary Ke and Simon Tong and Alex Fabrikant and William Bono and Jindong Gu and Ryan Foley and Yiran Mao and Manolis Delakis and Dhruva Bhaswar and Roy Frostig and Nick Li and Avital Zipori and Cath Hope and Olga Kozlova and Swaroop Mishra and Josip Djolonga and Craig Schiff and Majd Al Merey and Eleftheria Briakou and Peter Morgan and Andy Wan and Avinatan Hassidim and RJ Skerry-Ryan and Kuntal Sengupta and Mary Jasarevic and Praveen Kallakuri and Paige Kunkle and Hannah Brennan and Tom Lieber and Hassan Mansoor and Julian Walker and Bing Zhang and Annie Xie and Goran Žužić and Adaeze Chukwuka and Alex Druinsky and Donghyun Cho and Rui Yao and Ferjad Naeem and Shiraz Butt and Eunyoung Kim and Zhipeng Jia and Mandy Jordan and Adam Lelkes and Mark Kurzeja and Sophie Wang and James Zhao and Andrew Over and Abhishek Chakladar and Marcel Prasetya and Neha Jha and Sriram Ganapathy and Yale Cong and Prakash Shroff and Carl Saroufim and Sobhan Miryoosefi and Mohamed Hammad and Tajwar Nasir and Weijuan Xi and Yang Gao and Young Maeng and Ben Hora and Chin-Yi Cheng and Parisa Haghani and Yoad Lewenberg and Caden Lu and Martin Matysiak and Naina Raisinghani and Huiyu Wang and Lexi Baugher and Rahul Sukthankar and Minh Giang and John Schultz and Noah Fiedel and Minmin Chen and Cheng-Chun Lee and Tapomay Dey and Hao Zheng and Shachi Paul and Celine Smith and Andy Ly and Yicheng Wang and Rishabh Bansal and Bartek Perz and Susanna Ricco and Stasha Blank and Vaishakh Keshava and Deepak Sharma and Marvin Chow and Kunal Lad and Komal Jalan and Simon Osindero and Craig Swanson and Jacob Scott and Anastasija Ilić and Xiaowei Li and Siddhartha Reddy Jonnalagadda and Afzal Shama Soudagar and Yan Xiong and Bat-Orgil Batsaikhan and Daniel Jarrett and Naveen Kumar and Maulik Shah and Matt Lawlor and Austin Waters and Mark Graham and Rhys May and Sabela Ramos and Sandra Lefdal and Zeynep Cankara and Nacho Cano and Brendan O'Donoghue and Jed Borovik and Frederick Liu and Jordan Grimstad and Mahmoud Alnahlawi and Katerina Tsihlas and Tom Hudson and Nikolai Grigorev and Yiling Jia and Terry Huang and Tobenna Peter Igwe and Sergei Lebedev and Xiaodan Tang and Igor Krivokon and Frankie Garcia and Melissa Tan and Eric Jia and Peter Stys and Shikhar Vashishth and Yu Liang and Balaji Venkatraman and Chenjie Gu and Anastasios Kementsietsidis and Chen Zhu and Junehyuk Jung and Yunfei Bai and Mohammad Javad Hosseini and Faruk Ahmed and Aditya Gupta and Xin Yuan and Shereen Ashraf and Shitij Nigam and Gautam Vasudevan and Pranjal Awasthi and Adi Mayrav Gilady and Zelda Mariet and Ramy Eskander and Haiguang Li and Hexiang Hu and Guillermo Garrido and Philippe Schlattner and George Zhang and Rohun Saxena and Petar Dević and Kritika Muralidharan and Ashwin Murthy and Yiqian Zhou and Min Choi and Arissa Wongpanich and Zhengdong Wang and Premal Shah and Yuntao Xu and Yiling Huang and Stephen Spencer and Alice Chen and James Cohan and Junjie Wang and Jonathan Tompson and Junru Wu and Ruba Haroun and Haiqiong Li and Blanca Huergo and Fan Yang and Tongxin Yin and James Wendt and Michael Bendersky and Rahma Chaabouni and Javier Snaider and Johan Ferret and Abhishek Jindal and Tara Thompson and Andrew Xue and Will Bishop and Shubham Milind Phal and Archit Sharma and Yunhsuan Sung and Prabakar Radhakrishnan and Mo Shomrat and Reeve Ingle and Roopali Vij and Justin Gilmer and Mihai Dorin Istin and Sam Sobell and Yang Lu and Emily Nottage and Dorsa Sadigh and Jeremiah Willcock and Tingnan Zhang and Steve Xu and Sasha Brown and Katherine Lee and Gary Wang and Yun Zhu and Yi Tay and Cheolmin Kim and Audrey Gutierrez and Abhanshu Sharma and Yongqin Xian and Sungyong Seo and Claire Cui and Elena Pochernina and Cip Baetu and Krzysztof Jastrzębski and Mimi Ly and Mohamed Elhawaty and Dan Suh and Eren Sezener and Pidong Wang and Nancy Yuen and George Tucker and Jiahao Cai and Zuguang Yang and Cindy Wang and Alex Muzio and Hai Qian and Jae Yoo and Derek Lockhart and Kevin R. McKee and Mandy Guo and Malika Mehrotra and Artur Mendonça and Sanket Vaibhav Mehta and Sherry Ben and Chetan Tekur and Jiaqi Mu and Muye Zhu and Victoria Krakovna and Hongrae Lee and AJ Maschinot and Sébastien Cevey and HyunJeong Choe and Aijun Bai and Hansa Srinivasan and Derek Gasaway and Nick Young and Patrick Siegler and Dan Holtmann-Rice and Vihari Piratla and Kate Baumli and Roey Yogev and Alex Hofer and Hado van Hasselt and Svetlana Grant and Yuri Chervonyi and David Silver and Andrew Hogue and Ayushi Agarwal and Kathie Wang and Preeti Singh and Four Flynn and Josh Lipschultz and Robert David and Lizzetth Bellot and Yao-Yuan Yang and Long Le and Filippo Graziano and Kate Olszewska and Kevin Hui and Akanksha Maurya and Nikos Parotsidis and Weijie Chen and Tayo Oguntebi and Joe Kelley and Anirudh Baddepudi and Johannes Mauerer and Gregory Shaw and Alex Siegman and Lin Yang and Shravya Shetty and Subhrajit Roy and Yunting Song and Wojciech Stokowiec and Ryan Burnell and Omkar Savant and Robert Busa-Fekete and Jin Miao and Samrat Ghosh and Liam MacDermed and Phillip Lippe and Mikhail Dektiarev and Zach Behrman and Fabian Mentzer and Kelvin Nguyen and Meng Wei and Siddharth Verma and Chris Knutsen and Sudeep Dasari and Zhipeng Yan and Petr Mitrichev and Xingyu Wang and Virat Shejwalkar and Jacob Austin and Srinivas Sunkara and Navneet Potti and Yan Virin and Christian Wright and Gaël Liu and Oriana Riva and Etienne Pot and Greg Kochanski and Quoc Le and Gargi Balasubramaniam and Arka Dhar and Yuguo Liao and Adam Bloniarz and Divyansh Shukla and Elizabeth Cole and Jong Lee and Sheng Zhang and Sushant Kafle and Siddharth Vashishtha and Parsa Mahmoudieh and Grace Chen and Raphael Hoffmann and Pranesh Srinivasan and Agustin Dal Lago and Yoav Ben Shalom and Zi Wang and Michael Elabd and Anuj Sharma and Junhyuk Oh and Suraj Kothawade and Maigo Le and Marianne Monteiro and Shentao Yang and Kaiz Alarakyia and Robert Geirhos and Diana Mincu and Håvard Garnes and Hayato Kobayashi and Soroosh Mariooryad and Kacper Krasowiak and Zhixin and Lai and Shibl Mourad and Mingqiu Wang and Fan Bu and Ophir Aharoni and Guanjie Chen and Abhimanyu Goyal and Vadim Zubov and Ankur Bapna and Elahe Dabir and Nisarg Kothari and Kay Lamerigts and Nicola De Cao and Jeremy Shar and Christopher Yew and Nitish Kulkarni and Dre Mahaarachchi and Mandar Joshi and Zhenhai Zhu and Jared Lichtarge and Yichao Zhou and Hannah Muckenhirn and Vittorio Selo and Oriol Vinyals and Peter Chen and Anthony Brohan and Vaibhav Mehta and Sarah Cogan and Ruth Wang and Ty Geri and Wei-Jen Ko and Wei Chen and Fabio Viola and Keshav Shivam and Lisa Wang and Madeleine Clare Elish and Raluca Ada Popa and Sébastien Pereira and Jianqiao Liu and Raphael Koster and Donnie Kim and Gufeng Zhang and Sayna Ebrahimi and Partha Talukdar and Yanyan Zheng and Petra Poklukar and Ales Mikhalap and Dale Johnson and Anitha Vijayakumar and Mark Omernick and Matt Dibb and Ayush Dubey and Qiong Hu and Apurv Suman and Vaibhav Aggarwal and Ilya Kornakov and Fei Xia and Wing Lowe and Alexey Kolganov and Ted Xiao and Vitaly Nikolaev and Steven Hemingray and Bonnie Li and Joana Iljazi and Mikołaj Rybiński and Ballie Sandhu and Peggy Lu and Thang Luong and Rodolphe Jenatton and Vineetha Govindaraj and Hui and Li and Gabriel Dulac-Arnold and Wonpyo Park and Henry Wang and Abhinit Modi and Jean Pouget-Abadie and Kristina Greller and Rahul Gupta and Robert Berry and Prajit Ramachandran and Jinyu Xie and Liam McCafferty and Jianling Wang and Kilol Gupta and Hyeontaek Lim and Blaž Bratanič and Andy Brock and Ilia Akolzin and Jim Sproch and Dan Karliner and Duhyeon Kim and Adrian Goedeckemeyer and Noam Shazeer and Cordelia Schmid and Daniele Calandriello and Parul Bhatia and Krzysztof Choromanski and Ceslee Montgomery and Dheeru Dua and Ana Ramalho and Helen King and Yue Gao and Lynn Nguyen and David Lindner and Divya Pitta and Oleaser Johnson and Khalid Salama and Diego Ardila and Michael Han and Erin Farnese and Seth Odoom and Ziyue Wang and Xiangzhuo Ding and Norman Rink and Ray Smith and Harshal Tushar Lehri and Eden Cohen and Neera Vats and Tong He and Parthasarathy Gopavarapu and Adam Paszke and Miteyan Patel and Wouter Van Gansbeke and Lucia Loher and Luis Castro and Maria Voitovich and Tamara von Glehn and Nelson George and Simon Niklaus and Zach Eaton-Rosen and Nemanja Rakićević and Erik Jue and Sagi Perel and Carrie Zhang and Yuval Bahat and Angéline Pouget and Zhi Xing and Fantine Huot and Ashish Shenoy and Taylor Bos and Vincent Coriou and Bryan Richter and Natasha Noy and Yaqing Wang and Santiago Ontanon and Siyang Qin and Gleb Makarchuk and Demis Hassabis and Zhuowan Li and Mandar Sharma and Kumaran Venkatesan and Iurii Kemaev and Roxanne Daniel and Shiyu Huang and Saloni Shah and Octavio Ponce and Warren and Chen and Manaal Faruqui and Jialin Wu and Slavica Andačić and Szabolcs Payrits and Daniel McDuff and Tom Hume and Yuan Cao and MH Tessler and Qingze Wang and Yinan Wang and Ivor Rendulic and Eirikur Agustsson and Matthew Johnson and Tanya Lando and Andrew Howard and Sri Gayatri Sundara Padmanabhan and Mayank Daswani and Andrea Banino and Michael Kilgore and Jonathan Heek and Ziwei Ji and Alvaro Caceres and Conglong Li and Nora Kassner and Alexey Vlaskin and Zeyu Liu and Alex Grills and Yanhan Hou and Roykrong Sukkerd and Gowoon Cheon and Nishita Shetty and Larisa Markeeva and Piotr Stanczyk and Tejas Iyer and Yuan Gong and Shawn Gao and Keerthana Gopalakrishnan and Tim Blyth and Malcolm Reynolds and Avishkar Bhoopchand and Misha Bilenko and Dero Gharibian and Vicky Zayats and Aleksandra Faust and Abhinav Singh and Min Ma and Hongyang Jiao and Sudheendra Vijayanarasimhan and Lora Aroyo and Vikas Yadav and Sarah Chakera and Ashwin Kakarla and Vilobh Meshram and Karol Gregor and Gabriela Botea and Evan Senter and Dawei Jia and Geza Kovacs and Neha Sharma and Sebastien Baur and Kai Kang and Yifan He and Lin Zhuo and Marija Kostelac and Itay Laish and Songyou Peng and Louis O'Bryan and Daniel Kasenberg and Girish Ramchandra Rao and Edouard Leurent and Biao Zhang and Sage Stevens and Ana Salazar and Ye Zhang and Ivan Lobov and Jake Walker and Allen Porter and Morgan Redshaw and Han Ke and Abhishek Rao and Alex Lee and Hoi Lam and Michael Moffitt and Jaeyoun Kim and Siyuan Qiao and Terry Koo and Robert Dadashi and Xinying Song and Mukund Sundararajan and Peng Xu and Chizu Kawamoto and Yan Zhong and Clara Barbu and Apoorv Reddy and Mauro Verzetti and Leon Li and George Papamakarios and Hanna Klimczak-Plucińska and Mary Cassin and Koray Kavukcuoglu and Rigel Swavely and Alain Vaucher and Jeffrey Zhao and Ross Hemsley and Michael Tschannen and Heming Ge and Gaurav Menghani and Yang Yu and Natalie Ha and Wei He and Xiao Wu and Maggie Song and Rachel Sterneck and Stefan Zinke and Dan A. Calian and Annie Marsden and Alejandro Cruzado Ruiz and Matteo Hessel and Almog Gueta and Benjamin Lee and Brian Farris and Manish Gupta and Yunjie Li and Mohammad Saleh and Vedant Misra and Kefan Xiao and Piermaria Mendolicchio and Gavin Buttimore and Varvara Krayvanova and Nigamaa Nayakanti and Matthew Wiethoff and Yash Pande and Azalia Mirhoseini and Ni Lao and Jasmine Liu and Yiqing Hua and Angie Chen and Yury Malkov and Dmitry Kalashnikov and Shubham Gupta and Kartik Audhkhasi and Yuexiang Zhai and Sudhindra Kopalle and Prateek Jain and Eran Ofek and Clemens Meyer and Khuslen Baatarsukh and Hana Strejček and Jun Qian and James Freedman and Ricardo Figueira and Michal Sokolik and Olivier Bachem and Raymond Lin and Dia Kharrat and Chris Hidey and Pingmei Xu and Dennis Duan and Yin Li and Muge Ersoy and Richard Everett and Kevin Cen and Rebeca Santamaria-Fernandez and Amir Taubenfeld and Ian Mackinnon and Linda Deng and Polina Zablotskaia and Shashank Viswanadha and Shivanker Goel and Damion Yates and Yunxiao Deng and Peter Choy and Mingqing Chen and Abhishek Sinha and Alex Mossin and Yiming Wang and Arthur Szlam and Susan Hao and Paul Kishan Rubenstein and Metin Toksoz-Exley and Miranda Aperghis and Yin Zhong and Junwhan Ahn and Michael Isard and Olivier Lacombe and Florian Luisier and Chrysovalantis Anastasiou and Yogesh Kalley and Utsav Prabhu and Emma Dunleavy and Shaan Bijwadia and Justin Mao-Jones and Kelly Chen and Rama Pasumarthi and Emily Wood and Adil Dostmohamed and Nate Hurley and Jiri Simsa and Alicia Parrish and Mantas Pajarskas and Matt Harvey and Ondrej Skopek and Yony Kochinski and Javier Rey and Verena Rieser and Denny Zhou and Sun Jae Lee and Trilok Acharya and Guowang Li and Joe Jiang and Xiaofan Zhang and Bryant Gipson and Ethan Mahintorabi and Marco Gelmi and Nima Khajehnouri and Angel Yeh and Kayi Lee and Loic Matthey and Leslie Baker and Trang Pham and Han Fu and Alex Pak and Prakhar Gupta and Cristina Vasconcelos and Adam Sadovsky and Brian Walker and Sissie Hsiao and Patrik Zochbauer and Andreea Marzoca and Noam Velan and Junhao Zeng and Gilles Baechler and Danny Driess and Divya Jain and Yanping Huang and Lizzie Tao and John Maggs and Nir Levine and Jon Schneider and Erika Gemzer and Samuel Petit and Shan Han and Zach Fisher and Dustin Zelle and Courtney Biles and Eugene Ie and Asya Fadeeva and Casper Liu and Juliana Vicente Franco and Adrian Collister and Hao Zhang and Renshen Wang and Ruizhe Zhao and Leandro Kieliger and Kurt Shuster and Rui Zhu and Boqing Gong and Lawrence Chan and Ruoxi Sun and Sujoy Basu and Roland Zimmermann and Jamie Hayes and Abhishek Bapna and Jasper Snoek and Weel Yang and Puranjay Datta and Jad Al Abdallah and Kevin Kilgour and Lu Li and SQ Mah and Yennie Jun and Morgane Rivière and Abhijit Karmarkar and Tammo Spalink and Tao Huang and Lucas Gonzalez and Duc-Hieu Tran and Averi Nowak and John Palowitch and Martin Chadwick and Ellie Talius and Harsh Mehta and Thibault Sellam and Philipp Fränken and Massimo Nicosia and Kyle He and Aditya Kini and David Amos and Sugato Basu and Harrison Jobe and Eleni Shaw and Qiantong Xu and Colin Evans and Daisuke Ikeda and Chaochao Yan and Larry Jin and Lun Wang and Sachin Yadav and Ilia Labzovsky and Ramesh Sampath and Ada Ma and Candice Schumann and Aditya Siddhant and Rohin Shah and John Youssef and Rishabh Agarwal and Natalie Dabney and Alessio Tonioni and Moran Ambar and Jing Li and Isabelle Guyon and Benny Li and David Soergel and Boya Fang and Georgi Karadzhov and Cristian Udrescu and Trieu Trinh and Vikas Raunak and Seb Noury and Dee Guo and Sonal Gupta and Mara Finkelstein and Denis Petek and Lihao Liang and Greg Billock and Pei Sun and David Wood and Yiwen Song and Xiaobin Yu and Tatiana Matejovicova and Regev Cohen and Kalyan Andra and David D'Ambrosio and Zhiwei Deng and Vincent Nallatamby and Ebrahim Songhori and Rumen Dangovski and Andrew Lampinen and Pankil Botadra and Adam Hillier and Jiawei Cao and Nagabhushan Baddi and Adhi Kuncoro and Toshihiro Yoshino and Ankit Bhagatwala and Marcáurelio Ranzato and Rylan Schaeffer and Tianlin Liu and Shuai Ye and Obaid Sarvana and John Nham and Chenkai Kuang and Isabel Gao and Jinoo Baek and Shubham Mittal and Ayzaan Wahid and Anita Gergely and Bin Ni and Josh Feldman and Carrie Muir and Pascal Lamblin and Wolfgang Macherey and Ethan Dyer and Logan Kilpatrick and Víctor Campos and Mukul Bhutani and Stanislav Fort and Yanif Ahmad and Aliaksei Severyn and Kleopatra Chatziprimou and Oleksandr Ferludin and Mason Dimarco and Aditya Kusupati and Joe Heyward and Dan Bahir and Kevin Villela and Katie Millican and Dror Marcus and Sanaz Bahargam and Caglar Unlu and Nicholas Roth and Zichuan Wei and Siddharth Gopal and Deepanway Ghoshal and Edward Lee and Sharon Lin and Jennie Lees and Dayeong Lee and Anahita Hosseini and Connie Fan and Seth Neel and Marcus Wu and Yasemin Altun and Honglong Cai and Enrique Piqueras and Josh Woodward and Alessandro Bissacco and Salem Haykal and Mahyar Bordbar and Prasha Sundaram and Sarah Hodkinson and Daniel Toyama and George Polovets and Austin Myers and Anu Sinha and Tomer Levinboim and Kashyap Krishnakumar and Rachita Chhaparia and Tatiana Sholokhova and Nitesh Bharadwaj Gundavarapu and Ganesh Jawahar and Haroon Qureshi and Jieru Hu and Nikola Momchev and Matthew Rahtz and Renjie Wu and Aishwarya P S and Kedar Dhamdhere and Meiqi Guo and Umang Gupta and Ali Eslami and Mariano Schain and Michiel Blokzijl and David Welling and Dave Orr and Levent Bolelli and Nicolas Perez-Nieves and Mikhail Sirotenko and Aman Prasad and Arjun Kar and Borja De Balle Pigem and Tayfun Terzi and Gellért Weisz and Dipankar Ghosh and Aditi Mavalankar and Dhruv Madeka and Kaspar Daugaard and Hartwig Adam and Viraj Shah and Dana Berman and Maggie Tran and Steven Baker and Ewa Andrejczuk and Grishma Chole and Ganna Raboshchuk and Mahdi Mirzazadeh and Thais Kagohara and Shimu Wu and Christian Schallhart and Bernett Orlando and Chen Wang and Alban Rrustemi and Hao Xiong and Hao Liu and Arpi Vezer and Nolan Ramsden and Shuo-yiin Chang and Sidharth Mudgal and Yan Li and Nino Vieillard and Yedid Hoshen and Farooq Ahmad and Ambrose Slone and Amy Hua and Natan Potikha and Mirko Rossini and Jon Stritar and Sushant Prakash and Zifeng Wang and Xuanyi Dong and Alireza Nazari and Efrat Nehoran and Kaan Tekelioglu and Yinxiao Li and Kartikeya Badola and Tom Funkhouser and Yuanzhen Li and Varun Yerram and Ramya Ganeshan and Daniel Formoso and Karol Langner and Tian Shi and Huijian Li and Yumeya Yamamori and Amayika Panda and Alaa Saade and Angelo Scorza Scarpati and Chris Breaux and CJ Carey and Zongwei Zhou and Cho-Jui Hsieh and Sophie Bridgers and Alena Butryna and Nishesh Gupta and Vaibhav Tulsyan and Sanghyun Woo and Evgenii Eltyshev and Will Grathwohl and Chanel Parks and Seth Benjamin and Rina Panigrahy and Shenil Dodhia and Daniel De Freitas and Chris Sauer and Will Song and Ferran Alet and Jackson Tolins and Cosmin Paduraru and Xingyi Zhou and Brian Albert and Zizhao Zhang and Lei Shu and Mudit Bansal and Sarah Nguyen and Amir Globerson and Owen Xiao and James Manyika and Tom Hennigan and Rong Rong and Josip Matak and Anton Bakalov and Ankur Sharma and Danila Sinopalnikov and Andrew Pierson and Stephen Roller and Geoff Brown and Mingcen Gao and Toshiyuki Fukuzawa and Amin Ghafouri and Kenny Vassigh and Iain Barr and Zhicheng Wang and Anna Korsun and Rajesh Jayaram and Lijie Ren and Tim Zaman and Samira Khan and Yana Lunts and Dan Deutsch and Dave Uthus and Nitzan Katz and Masha Samsikova and Amr Khalifa and Nikhil Sethi and Jiao Sun and Luming Tang and Uri Alon and Xianghong Luo and Dian Yu and Abhishek Nayyar and Bryce Petrini and Will Truong and Vincent Hellendoorn and Nikolai Chinaev and Chris Alberti and Wei Wang and Jingcao Hu and Vahab Mirrokni and Ananth Balashankar and Avia Aharon and Aahil Mehta and Ahmet Iscen and Joseph Kready and Lucas Manning and Anhad Mohananey and Yuankai Chen and Anshuman Tripathi and Allen Wu and Igor Petrovski and Dawsen Hwang and Martin Baeuml and Shreyas Chandrakaladharan and Yuan Liu and Rey Coaguila and Maxwell Chen and Sally Ma and Pouya Tafti and Susheel Tatineni and Terry Spitz and Jiayu Ye and Paul Vicol and Mihaela Rosca and Adrià Puigdomènech and Zohar Yahav and Sanjay Ghemawat and Hanzhao Lin and Phoebe Kirk and Zaid Nabulsi and Sergey Brin and Bernd Bohnet and Ken Caluwaerts and Aditya Srikanth Veerubhotla and Dan Zheng and Zihang Dai and Petre Petrov and Yichong Xu and Ramin Mehran and Zhuo Xu and Luisa Zintgraf and Jiho Choi and Spurthi Amba Hombaiah and Romal Thoppilan and Sashank Reddi and Lukasz Lew and Li Li and Kellie Webster and KP Sawhney and Lampros Lamprou and Siamak Shakeri and Mayank Lunayach and Jianmin Chen and Sumit Bagri and Alex Salcianu and Ying Chen and Yani Donchev and Charlotte Magister and Signe Nørly and Vitor Rodrigues and Tomas Izo and Hila Noga and Joe Zou and Thomas Köppe and Wenxuan Zhou and Kenton Lee and Xiangzhu Long and Danielle Eisenbud and Anthony Chen and Connor Schenck and Chi Ming To and Peilin Zhong and Emanuel Taropa and Minh Truong and Omer Levy and Danilo Martins and Zhiyuan Zhang and Christopher Semturs and Kelvin Zhang and Alex Yakubovich and Pol Moreno and Lara McConnaughey and Di Lu and Sam Redmond and Lotte Weerts and Yonatan Bitton and Tiziana Refice and Nicolas Lacasse and Arthur Conmy and Corentin Tallec and Julian Odell and Hannah Forbes-Pollard and Arkadiusz Socala and Jonathan Hoech and Pushmeet Kohli and Alanna Walton and Rui Wang and Mikita Sazanovich and Kexin Zhu and Andrei Kapishnikov and Rich Galt and Matthew Denton and Ben Murdoch and Caitlin Sikora and Kareem Mohamed and Wei Wei and Uri First and Tim McConnell and Luis C. Cobo and James Qin and Thi Avrahami and Daniel Balle and Yu Watanabe and Annie Louis and Adam Kraft and Setareh Ariafar and Yiming Gu and Eugénie Rives and Charles Yoon and Andrei Rusu and James Cobon-Kerr and Chris Hahn and Jiaming Luo and Yuvein and Zhu and Niharika Ahuja and Rodrigo Benenson and Raphaël Lopez Kaufman and Honglin Yu and Lloyd Hightower and Junlin Zhang and Darren Ni and Lisa Anne Hendricks and Gabby Wang and Gal Yona and Lalit Jain and Pablo Barrio and Surya Bhupatiraju and Siva Velusamy and Allan Dafoe and Sebastian Riedel and Tara Thomas and Zhe Yuan and Mathias Bellaiche and Sheena Panthaplackel and Klemen Kloboves and Sarthak Jauhari and Canfer Akbulut and Todor Davchev and Evgeny Gladchenko and David Madras and Aleksandr Chuklin and Tyrone Hill and Quan Yuan and Mukundan Madhavan and Luke Leonhard and Dylan Scandinaro and Qihang Chen and Ning Niu and Arthur Douillard and Bogdan Damoc and Yasumasa Onoe and Fabian Pedregosa and Fred Bertsch and Chas Leichner and Joseph Pagadora and Jonathan Malmaud and Sameera Ponda and Andy Twigg and Oleksii Duzhyi and Jingwei Shen and Miaosen Wang and Roopal Garg and Jing Chen and Utku Evci and Jonathan Lee and Leon Liu and Koji Kojima and Masa Yamaguchi and Arunkumar Rajendran and AJ Piergiovanni and Vinodh Kumar Rajendran and Marco Fornoni and Gabriel Ibagon and Harry Ragan and Sadh MNM Khan and John Blitzer and Andrew Bunner and Guan Sun and Takahiro Kosakai and Scott Lundberg and Ndidi Elue and Kelvin Guu and SK Park and Jane Park and Arunachalam Narayanaswamy and Chengda Wu and Jayaram Mudigonda and Trevor Cohn and Hairong Mu and Ravi Kumar and Laura Graesser and Yichi Zhang and Richard Killam and Vincent Zhuang and Mai Giménez and Wael Al Jishi and Ruy Ley-Wild and Alex Zhai and Kazuki Osawa and Diego Cedillo and Jialu Liu and Mayank Upadhyay and Marcin Sieniek and Roshan Sharma and Tom Paine and Anelia Angelova and Sravanti Addepalli and Carolina Parada and Kingshuk Majumder and Avery Lamp and Sanjiv Kumar and Xiang Deng and Artiom Myaskovsky and Tea Sabolić and Jeffrey Dudek and Sarah York and Félix de Chaumont Quitry and Jiazhong Nie and Dee Cattle and Alok Gunjan and Bilal Piot and Waleed Khawaja and Seojin Bang and Simon Wang and Siavash Khodadadeh and Raghavender R and Praynaa Rawlani and Richard Powell and Kevin Lee and Johannes Griesser and GS Oh and Cesar Magalhaes and Yujia Li and Simon Tokumine and Hadas Natalie Vogel and Dennis Hsu and Arturo BC and Disha Jindal and Matan Cohen and Zi Yang and Junwei Yuan and Dario de Cesare and Tony Bruguier and Jun Xu and Monica Roy and Alon Jacovi and Dan Belov and Rahul Arya and Phoenix Meadowlark and Shlomi Cohen-Ganor and Wenting Ye and Patrick Morris-Suzuki and Praseem Banzal and Gan Song and Pranavaraj Ponnuramu and Fred Zhang and George Scrivener and Salah Zaiem and Alif Raditya Rochman and Kehang Han and Badih Ghazi and Kate Lee and Shahar Drath and Daniel Suo and Antonious Girgis and Pradeep Shenoy and Duy Nguyen and Douglas Eck and Somit Gupta and Le Yan and Joao Carreira and Anmol Gulati and Ruoxin Sang and Daniil Mirylenka and Emma Cooney and Edward Chou and Mingyang Ling and Cindy Fan and Ben Coleman and Guilherme Tubone and Ravin Kumar and Jason Baldridge and Felix Hernandez-Campos and Angeliki Lazaridou and James Besley and Itay Yona and Neslihan Bulut and Quentin Wellens and AJ Pierigiovanni and Jasmine George and Richard Green and Pu Han and Connie Tao and Geoff Clark and Chong You and Abbas Abdolmaleki and Justin Fu and Tongzhou Chen and Ashwin Chaugule and Angad Chandorkar and Altaf Rahman and Will Thompson and Penporn Koanantakool and Mike Bernico and Jie Ren and Andrey Vlasov and Sergei Vassilvitskii and Maciej Kula and Yizhong Liang and Dahun Kim and Yangsibo Huang and Chengxi Ye and Dmitry Lepikhin and Wesley Helmholz},
      year={2025},
      eprint={2507.06261},
      archivePrefix={arXiv},
      primaryClass={cs.CL},
      url={https://arxiv.org/abs/2507.06261}, 
}

@misc{yang2025qwen3technicalreport,
      title={Qwen3 Technical Report}, 
      author={An Yang and Anfeng Li and Baosong Yang and Beichen Zhang and Binyuan Hui and Bo Zheng and Bowen Yu and Chang Gao and Chengen Huang and Chenxu Lv and Chujie Zheng and Dayiheng Liu and Fan Zhou and Fei Huang and Feng Hu and Hao Ge and Haoran Wei and Huan Lin and Jialong Tang and Jian Yang and Jianhong Tu and Jianwei Zhang and Jianxin Yang and Jiaxi Yang and Jing Zhou and Jingren Zhou and Junyang Lin and Kai Dang and Keqin Bao and Kexin Yang and Le Yu and Lianghao Deng and Mei Li and Mingfeng Xue and Mingze Li and Pei Zhang and Peng Wang and Qin Zhu and Rui Men and Ruize Gao and Shixuan Liu and Shuang Luo and Tianhao Li and Tianyi Tang and Wenbiao Yin and Xingzhang Ren and Xinyu Wang and Xinyu Zhang and Xuancheng Ren and Yang Fan and Yang Su and Yichang Zhang and Yinger Zhang and Yu Wan and Yuqiong Liu and Zekun Wang and Zeyu Cui and Zhenru Zhang and Zhipeng Zhou and Zihan Qiu},
      year={2025},
      eprint={2505.09388},
      archivePrefix={arXiv},
      primaryClass={cs.CL},
      url={https://arxiv.org/abs/2505.09388}, 
}

@misc{gemmateam2025gemma3technicalreport,
      title={Gemma 3 Technical Report}, 
      author={Gemma Team and Aishwarya Kamath and Johan Ferret and Shreya Pathak and Nino Vieillard and Ramona Merhej and Sarah Perrin and Tatiana Matejovicova and Alexandre Ramé and Morgane Rivière and Louis Rouillard and Thomas Mesnard and Geoffrey Cideron and Jean-bastien Grill and Sabela Ramos and Edouard Yvinec and Michelle Casbon and Etienne Pot and Ivo Penchev and Gaël Liu and Francesco Visin and Kathleen Kenealy and Lucas Beyer and Xiaohai Zhai and Anton Tsitsulin and Robert Busa-Fekete and Alex Feng and Noveen Sachdeva and Benjamin Coleman and Yi Gao and Basil Mustafa and Iain Barr and Emilio Parisotto and David Tian and Matan Eyal and Colin Cherry and Jan-Thorsten Peter and Danila Sinopalnikov and Surya Bhupatiraju and Rishabh Agarwal and Mehran Kazemi and Dan Malkin and Ravin Kumar and David Vilar and Idan Brusilovsky and Jiaming Luo and Andreas Steiner and Abe Friesen and Abhanshu Sharma and Abheesht Sharma and Adi Mayrav Gilady and Adrian Goedeckemeyer and Alaa Saade and Alex Feng and Alexander Kolesnikov and Alexei Bendebury and Alvin Abdagic and Amit Vadi and András György and André Susano Pinto and Anil Das and Ankur Bapna and Antoine Miech and Antoine Yang and Antonia Paterson and Ashish Shenoy and Ayan Chakrabarti and Bilal Piot and Bo Wu and Bobak Shahriari and Bryce Petrini and Charlie Chen and Charline Le Lan and Christopher A. Choquette-Choo and CJ Carey and Cormac Brick and Daniel Deutsch and Danielle Eisenbud and Dee Cattle and Derek Cheng and Dimitris Paparas and Divyashree Shivakumar Sreepathihalli and Doug Reid and Dustin Tran and Dustin Zelle and Eric Noland and Erwin Huizenga and Eugene Kharitonov and Frederick Liu and Gagik Amirkhanyan and Glenn Cameron and Hadi Hashemi and Hanna Klimczak-Plucińska and Harman Singh and Harsh Mehta and Harshal Tushar Lehri and Hussein Hazimeh and Ian Ballantyne and Idan Szpektor and Ivan Nardini and Jean Pouget-Abadie and Jetha Chan and Joe Stanton and John Wieting and Jonathan Lai and Jordi Orbay and Joseph Fernandez and Josh Newlan and Ju-yeong Ji and Jyotinder Singh and Kat Black and Kathy Yu and Kevin Hui and Kiran Vodrahalli and Klaus Greff and Linhai Qiu and Marcella Valentine and Marina Coelho and Marvin Ritter and Matt Hoffman and Matthew Watson and Mayank Chaturvedi and Michael Moynihan and Min Ma and Nabila Babar and Natasha Noy and Nathan Byrd and Nick Roy and Nikola Momchev and Nilay Chauhan and Noveen Sachdeva and Oskar Bunyan and Pankil Botarda and Paul Caron and Paul Kishan Rubenstein and Phil Culliton and Philipp Schmid and Pier Giuseppe Sessa and Pingmei Xu and Piotr Stanczyk and Pouya Tafti and Rakesh Shivanna and Renjie Wu and Renke Pan and Reza Rokni and Rob Willoughby and Rohith Vallu and Ryan Mullins and Sammy Jerome and Sara Smoot and Sertan Girgin and Shariq Iqbal and Shashir Reddy and Shruti Sheth and Siim Põder and Sijal Bhatnagar and Sindhu Raghuram Panyam and Sivan Eiger and Susan Zhang and Tianqi Liu and Trevor Yacovone and Tyler Liechty and Uday Kalra and Utku Evci and Vedant Misra and Vincent Roseberry and Vlad Feinberg and Vlad Kolesnikov and Woohyun Han and Woosuk Kwon and Xi Chen and Yinlam Chow and Yuvein Zhu and Zichuan Wei and Zoltan Egyed and Victor Cotruta and Minh Giang and Phoebe Kirk and Anand Rao and Kat Black and Nabila Babar and Jessica Lo and Erica Moreira and Luiz Gustavo Martins and Omar Sanseviero and Lucas Gonzalez and Zach Gleicher and Tris Warkentin and Vahab Mirrokni and Evan Senter and Eli Collins and Joelle Barral and Zoubin Ghahramani and Raia Hadsell and Yossi Matias and D. Sculley and Slav Petrov and Noah Fiedel and Noam Shazeer and Oriol Vinyals and Jeff Dean and Demis Hassabis and Koray Kavukcuoglu and Clement Farabet and Elena Buchatskaya and Jean-Baptiste Alayrac and Rohan Anil and Dmitry and Lepikhin and Sebastian Borgeaud and Olivier Bachem and Armand Joulin and Alek Andreev and Cassidy Hardin and Robert Dadashi and Léonard Hussenot},
      year={2025},
      eprint={2503.19786},
      archivePrefix={arXiv},
      primaryClass={cs.CL},
      url={https://arxiv.org/abs/2503.19786}, 
}

@article{kim2025fine,
  title={Fine-grained and thematic evaluation of llms in social deduction game},
  author={Kim, Byungjun and Seo, Dayeon and Kim, Minju and Kim, Bugeun},
  journal={IEEE Access},
  year={2025},
  publisher={IEEE}
}

@misc{song2025survivalevaluatingllmssocial,
      title={Beyond Survival: Evaluating LLMs in Social Deduction Games with Human-Aligned Strategies}, 
      author={Zirui Song and Yuan Huang and Junchang Liu and Haozhe Luo and Chenxi Wang and Lang Gao and Zixiang Xu and Mingfei Han and Xiaojun Chang and Xiuying Chen},
      year={2025},
      eprint={2510.11389},
      archivePrefix={arXiv},
      primaryClass={cs.CL},
      url={https://arxiv.org/abs/2510.11389}, 
}

@inproceedings{hugamearena,
  title={GameArena: Evaluating LLM Reasoning through Live Computer Games},
  author={Hu, Lanxiang and Li, Qiyu and Xie, Anze and Jiang, Nan and Stoica, Ion and Jin, Haojian and Zhang, Hao},
  booktitle={The Thirteenth International Conference on Learning Representations}
}

@inproceedings{hu2020xtreme,
  title={Xtreme: A massively multilingual multi-task benchmark for evaluating cross-lingual generalisation},
  author={Hu, Junjie and Ruder, Sebastian and Siddhant, Aditya and Neubig, Graham and Firat, Orhan and Johnson, Melvin},
  booktitle={International conference on machine learning},
  pages={4411--4421},
  year={2020},
  organization={PMLR}
}

@inproceedings{moniri-etal-2025-debate,
    title = "Evaluating the Performance of Large Language Models via Debates",
    author = "Moniri, Behrad  and
      Hassani, Hamed  and
      Dobriban, Edgar",
    editor = "Chiruzzo, Luis  and
      Ritter, Alan  and
      Wang, Lu",
    booktitle = "Findings of the Association for Computational Linguistics: NAACL 2025",
    month = apr,
    year = "2025",
    address = "Albuquerque, New Mexico",
    publisher = "Association for Computational Linguistics",
    url = "https://aclanthology.org/2025.findings-naacl.109/",
    doi = "10.18653/v1/2025.findings-naacl.109",
    pages = "2040--2075",
    ISBN = "979-8-89176-195-7"
}

@misc{wu2025bitterlessonlearned2000,
      title={The Bitter Lesson Learned from 2,000+ Multilingual Benchmarks}, 
      author={Minghao Wu and Weixuan Wang and Sinuo Liu and Huifeng Yin and Xintong Wang and Yu Zhao and Chenyang Lyu and Longyue Wang and Weihua Luo and Kaifu Zhang},
      year={2025},
      eprint={2504.15521},
      archivePrefix={arXiv},
      primaryClass={cs.CL},
      url={https://arxiv.org/abs/2504.15521}, 
}

\appendix
\newpage

\section{Selected Entities} \label{appendix:entities}

The entities that we selected with care for the model for each scenario:

\noindent \textbf{Generic Entities (EN):} Airplane, Amusement Park, Bank, Beach, Carnival, Casino, Circus Tent, Corporate Party, Crusader Army, Day Spa, Embassy, Hospital, Hotel, Military Base, Movie Studio, Nightclub, Ocean Liner, Passenger Train, Police Station, Pirate Ship, Polar Station, Restaurant, School, Service Station, Space Station, Submarine, Supermarket, Theater, University, Zoo.

\noindent \textbf{Generic Entities (ID):} Pesawat Terbang, Taman Hiburan, Bank, Pantai, Karnaval, Kasino, Tenda Sirkus, Pesta Perusahaan, Pasukan Perang Salib, Spa, Kedutaan Besar, Rumah Sakit, Hotel, Pangkalan Militer, Studio Film, Klub Malam, Kapal Pesiar, Kereta Penumpang, Kantor Polisi, Kapal Bajak Laut, Stasiun Kutub, Restoran, Sekolah, Bengkel, Stasiun Luar Angkasa, Kapal Selam, Supermarket, Teater, Universitas, Kebun Binatang.

\noindent \textbf{Generic Entities (ZH):} \zh{飞机}, \zh{游乐园}, \zh{银行}, \zh{海滩}, \zh{嘉年华}, \zh{赌场}, \zh{马戏团帐篷}, \zh{公司派对}, \zh{十字军}, \zh{水疗中心}, \zh{大使馆}, \zh{医院}, \zh{酒店}, \zh{军事基地}, \zh{电影制片厂}, \zh{夜总会}, \zh{远洋客轮}, \zh{客运火车}, \zh{警察局}, \zh{海盗船}, \zh{极地站}, \zh{餐厅}, \zh{学校}, \zh{加油站}, \zh{空间站}, \zh{潜水艇}, \zh{超市}, \zh{剧院}, \zh{大学}, \zh{动物园}.

\noindent \textbf{Generic Entities (ARZ):} \arabictext{طيارة}, \arabictext{ملاهي}, \arabictext{بنك}, \arabictext{شاطئ}, \arabictext{كرنفال}, \arabictext{كازينو}, \arabictext{سيرك}, \arabictext{حفلة شركة}, \arabictext{جيش الصليبيين}, \arabictext{سبا نهاري}, \arabictext{سفارة}, \arabictext{مستشفى}, \arabictext{فندق}, \arabictext{قاعدة عسكرية}, \arabictext{ستوديو تصوير افلام}, \arabictext{نايت كلوب}, \arabictext{سفينة ركاب}, \arabictext{قطر ركاب}, \arabictext{قسم شرطة}, \arabictext{سفينة قراصنة}, \arabictext{محطة قطبية}, \arabictext{مطعم}, \arabictext{مدرسة}, \arabictext{محطة بنزين}, \arabictext{محطة فضاء}, \arabictext{غواصة}, \arabictext{سوبر ماركت}, \arabictext{مسرح}, \arabictext{جامعة}, \arabictext{حديقة الحيوان}.

\noindent \textbf{ARZ Food:} \arabictext{حمام محشي}, \arabictext{كحك العيد}, \arabictext{كراويه}, \arabictext{حلبة حصى}, \arabictext{كنافة}, \arabictext{شكشوكة}, \arabictext{ممبار}, \arabictext{بيض بالبسطرمة}, \arabictext{ملوخية}, \arabictext{ينسون}, \arabictext{حواوشي}, \arabictext{شوربة كوارع}, \arabictext{مسقعة}, \arabictext{حمص الشام}, \arabictext{محشي}, \arabictext{صينية بطاطس بالفراخ}, \arabictext{كباب حلة}, \arabictext{بيتي فور}, \arabictext{شوربة لسان عصفور}, \arabictext{كشري}, \arabictext{ترمس}, \arabictext{سحلب}, \arabictext{بطاطا مشوية}, \arabictext{صوابع زينب}, \arabictext{فول مدمس}, \arabictext{كركديه}, \arabictext{سوبيا}, \arabictext{فتة لحمة}, \arabictext{رز بلبن}, \arabictext{فسيخ ورنجة}.

\noindent \textbf{ID Food:} Nastar, Nasi Tumpeng, Roti Buaya, Nasi Uduk, Lontong Sayur, Rendang, Ayam Taliwang, Babi Guling, Gado Gado, Tempe Mendoan, Capcai, Martabak, Lapis Legit, Bika Ambon, Cimol, Sate, Pempek, Bakso, Coto Makasar, Rawon, Seblak, Cakalang, fufu rica-rica, Tuak, Cap Tikus, Cendol, Soda Gembira, Es Lidah Buaya, Beras Kencur, Bandrek, Loloh Cemcem.

\red{\noindent \textbf{ZH Food:} \zh{馒头}, \zh{花卷}, \zh{发糕}, \zh{窝头}, \zh{烧饼}, \zh{馕}, \zh{元宵}, \zh{粽子}, \zh{年糕}, \zh{鱼香肉丝}, \zh{回锅肉}, \zh{地三鲜}, \zh{水煮鱼}, \zh{辣子鸡}, \zh{白切鸡}, \zh{烧鹅}, \zh{羊肉泡馍}, \zh{葱爆羊肉}, \zh{东坡肉}, \zh{煎饼果子}, \zh{肉夹馍}, \zh{臭豆腐}, \zh{凉皮}, \zh{手抓饼}, \zh{串串香}, \zh{茶叶蛋}, \zh{肠粉}, \zh{双皮奶}, \zh{芝麻糊}, \zh{切糕}.}

\noindent \textbf{ARZ Places:} \arabictext{الجامعة الامريكية بالقاهرة}, \arabictext{المتحف المصري الكبير}, \arabictext{الاهرامات}, \arabictext{كايرو فيستيفال سيتي مول}, \arabictext{تسيباس}, \arabictext{كوبري ستانلي}, \arabictext{بور فؤاد}, \arabictext{حجوجة}, \arabictext{مسجد محمد علي}, \arabictext{جامع الازهر}, \arabictext{الكنيسة المعلقة}, \arabictext{جامعة النيل}, \arabictext{سكي مصر}, \arabictext{سان ستيفانو}, \arabictext{المتحف القومي للحضارة}, \arabictext{جامعة عين شمس}, \arabictext{برج القاهرة}, \arabictext{كوبري قصر النيل}, \arabictext{الحسين}, \arabictext{العبد}, \arabictext{مدينة نصر}, \arabictext{سيلانترو}, \arabictext{وادي الملوك}, \arabictext{حلوان}, \arabictext{سيتي ستارز}, \arabictext{دريم بارك}, \arabictext{جاد}, \arabictext{كشري ابو طارق}, \arabictext{مكتبة اسكندرية}, \arabictext{كبابجي الدهان}.

\noindent \textbf{ID Places:} Trans Studio, Malioboro, Ragunan, Tanah Abang, Menteng, J.CO, Pizza Domino, Fort Rotterdam, Ciater, Jam Gadang, Nusa Penida, Bromo, Dufan, Gedung Sate, ITB, MAN Insan Cendekia, SMAN 8 Jakarta, Binus, Tuku, Sushi Tei, Restoran Sederhana, Warteg, Taman Nasional Sebangau, Pura Agung Besakih, Monas, BSD City, Blok M, Istiqlal, Borobudur, Kota Tua.

\noindent \textbf{ZH Places:} \zh{瑞幸咖啡}, \zh{喜茶}, \zh{蜜雪冰城}, \zh{海底捞火锅}, \zh{西贝莜面村}, \zh{稻香村}, \zh{必胜客}, \zh{麦当劳}, \zh{王府井大街}, \zh{三里屯太古里}, \zh{南京路步行街}, \zh{故宫博物院}, \zh{秦始皇兵马俑}, \zh{上海迪士尼乐园}, \zh{天安门广场}, \zh{东方明珠电视塔}, \zh{布达拉宫}, \zh{卢沟桥}, \zh{乐山大佛}, \zh{都江堰}, \zh{北京大学}, \zh{清华大学}, \zh{复旦大学}, \zh{中国科学院}, \zh{上海交大}, \zh{少林寺}, \zh{雍和宫}, \zh{上海外滩}, \zh{陆家嘴}, \zh{中关村}.

\section{Prompt Templates} \label{appendix:prompt}

During each game phase, every model receives a prompt containing the rules, game history, a phase description, and response instructions. Responses must adhere to a strict JSON format; any output that fails to do so is deemed invalid. In such instances, the team responsible for the invalid move immediately forfeits the game \texttt{quit} (e.g., an invalid move by the spy results in a victory for the non-spies). We argue that a model's inability to follow structural instructions is a clear indicator of limited capability, making a loss a fair and representative outcome. The full prompts are provided in Appendix .

This is the rule prompt that is always inserted in the LLM's prompt, where \texttt{the languages} are: ``Indonesian'', ``Simplified Chinese'', ``Egyptian Arabic'', and ``English'' (depends on the scenario run). Each phase and role will have their respective prompt.  Here are the respective prompts:

\begin{lstlisting}[style=yamlstyle,basicstyle=\ttfamily\footnotesize,breaklines=true]
# Base context used for all prompts
base_context: |
  **Game Context:** You are playing "Spyfall: Structured Edition" with other players.
  
  **How Spyfall Structured Edition Works:**
  - One player is secretly assigned as "the Spy" at the start
  - All other players are "Villagers/Non-Spies" who know the secret entity
  - The Spy does not know the entity and must figure it out
  - Villagers must identify who the Spy is through questioning and voting
  
  **What is the Entity?**
  The entity can be anything: a location (Beach, Hospital), a movie (Star Wars, Titanic), 
  a famous person (Einstein, Beyoncé), an object (Smartphone, Pizza), a concept (Birthday Party, Job Interview), etc.
  Villagers know the entity; the Spy must deduce it from questions and answers.
  
  **Structured Game Flow (Predictable Phases):**
  
  1. Round Robin Phase (Phase 1A): 
     - Each player asks the next player one question sequentially
     - Goes around the full circle once
     - Everyone participates, building baseline information
  
  2. Free Question Cycles (n cycles of Phase 1B-2-3):
     Each cycle has three phases:
     
     a) Phase 1B - Free Question:
        - Last answerer asks anyone one question
        - Provides targeted interrogation opportunities
     
     b) Phase 2 - Spy Guess Decision:
        - Spy can guess the entity or skip
        - If correct guess → Spy wins
        - If incorrect guess → Villagers win
        - Spy gets n chances throughout the game
     
     c) Phase 3 - Accusation Vote:
        - All players vote for someone (or skip)
        - Need majority (>50%) to eliminate someone
        - If eliminated player is Spy → Villagers win
        - If eliminated player is Villager → Spy wins
        - No majority → game continues to next cycle
  
  3. Final Round (after n cycles):
     - Final Spy guess opportunity
     - Final accusation vote
     - If no majority in final vote → Spy wins (survived!)
  
  Note: n equals the number of players in the game.

  **Win Conditions:**
  - Villagers win: Spy guesses wrong OR majority votes eliminate Spy
  - Spy wins: Spy guesses correctly OR survives all votes OR Villagers eliminate innocent player
  
  **Strategic Differences from Classic:**
  - Voting happens in cycles, not randomly
  - Spy has multiple guess opportunities (strategic timing!)
  - Vote history is public (builds narrative across rounds)
  - Structured phases allow better planning
  
  **Your Task:** Based on your role (Spy or Villager), play strategically:
  
  *As a Villager:*
  Share entity knowledge subtly and build your case across voting rounds. Avoid being obvious! Be extremely subtle in giving hints. Think carefully about how you answer and ask questions strategically.
  
  Important guidelines:
  - Do not ask or answer questions too directly
  - Review the entity list carefully to avoid making it easy for the Spy to guess
  - Avoid narrowing it down to one specific place or category
  - Do not provide direct hints to the entity (e.g., avoid saying "crew" or "vessel" when the entity is Pirate Ship, or "extreme" when it's Polar Bear)
  - Use subtlety and misdirection to make the Spy think multiple entities are possible
  - Poor strategy example: Mentioning "patient" when the entity is Hospital, or "temporary stay" when the entity is Hotel
  - Be very subtle so the Spy still needs to guess among many possibilities
  
  *As a Spy:*
  - Blend in, gather clues across multiple rounds, and time your guess strategically. 
  - Avoid being obvious and don't reveal that you're the Spy. 
  - Answer strategically using misdirection and subtlety.
  - Do not guess the entity if you cannot narrow it down to one location yet in Spy Guess Decision phase.
  
  Important guidelines:
  - Blend in and avoid obvious behavior
  - Use misdirection and subtlety in your questions and answers
  - Poor strategy example: Asking "Is it a place where people go to relax?" when the entity is Beach
  - Be very subtle in your approach
  
  **Critical Reminders:**
  - Check the entity guesses list that has been provided to you
  - Use it to guide your questioning and answering strategy (for both Spy and Villagers)
  - Consider conversation history, voting patterns, and strategic phase timing
  - Ensure you adhere to the response format
  - You must speak in {language} language at any cost!

# Prompts for question generation
question_generation:
  spy: |
    {base_context}

    You are the Spy and need to ask a question to another player.

    **Response Format Requirements:**
    You must respond with only valid JSON wrapped in triple pipes |||...|||
    Think through your approach before outputting JSON. No text should appear after the JSON.
    Think step by step!

    Example:
    Let's think step by step...
    |||
    {
      "question": "YOUR QUESTION HERE",
      "targeted_player": "TARGET_PLAYER_NAME" 
    }
    |||

    Required fields:
    - question: string (the actual question to ask)
    - targeted_player: string (the player being targeted by the question. Think strategically about who to ask)

  non_spy: |
    {base_context}

    You are a Villager and need to ask a question.
    
    **Response Format Requirements:**
    You must respond with only valid JSON wrapped in triple pipes |||...|||
    Think through your approach before outputting JSON. No text should appear after the JSON.
    Think step by step!

    Example:
    Let's think step by step...
    |||
    {
      "question": "YOUR QUESTION HERE",
      "targeted_player": "TARGET_PLAYER_NAME"
    }
    |||

    Required fields:
    - question: string (the actual question to ask)
    - targeted_player: string (the player being targeted by the question. Think strategically about who to ask)

# Prompts for answer generation
answer_generation:
  spy: |
    {base_context}

    You are the Spy answering a question.

    **Response Format Requirements:**
    You must respond with only valid JSON wrapped in triple pipes |||...|||
    Think through your approach before outputting JSON. No text should appear after the JSON.
    Think step by step!

    Example:
    Let's think step by step...
    |||
    {
      "answer": "It can get a bit uncomfortable, so I usually prepare accordingly."
    }
    |||

    Required fields:
    - answer: string (your actual answer)

    Important: Do not make it obvious that you are the Spy!
    
  non_spy: |
    {base_context}

    You are a Villager answering a question.
  
    **Response Format Requirements:**
    You must respond with only valid JSON wrapped in triple pipes |||...|||
    Think through your approach before outputting JSON. No text should appear after the JSON.
    Think step by step!

    Example:
    Let's think step by step...
    |||
    {
      "answer": "YOUR ANSWER HERE"
    }
    |||

    Required fields:
    - answer: string (your actual answer)

# Prompts for entity guessing (spy only)
entity_guess:
  spy: |
    {base_context}

    You are the Spy in Phase 2 (Spy Guess Decision).

    Important: You do not have to rush if you are unsure! Skip if you haven't pinpointed a single location yet.
    
    Your task: Decide whether to guess the entity now or skip.

    **Response Format Requirements:**
    You must respond with only valid JSON wrapped in triple pipes |||...|||
    Think through your approach before outputting JSON. No text should appear after the JSON.
    Think step by step! 

    Example (making a guess):
    Let's think step by step...
    |||
    {
      "best_guess": "Beach",
      "should_guess": true,
      "confidence": 0.85
    }
    |||

    Example (skipping):
    Let's think step by step...
    |||
    {
      "best_guess": null,
      "should_guess": false,
      "confidence": 0.3
    }
    |||

    Required fields:
    - best_guess: string or null (your guess from the entity list if should_guess=true, else null)
    - should_guess: boolean (true to guess now, false to skip)
    - confidence: number 0.0-1.0 (how confident you are in your guess)

# Prompts for vote_initiation (deciding who to vote for in Structured Edition)
vote_initiation:
  spy: |
    {base_context}

    You are the Spy deciding who to vote for in the Accusation Vote phase.
    You can skip voting if you are unsure!

    **Response Format Requirements:**
    You must respond with only valid JSON wrapped in triple pipes |||...|||
    Think through your approach before outputting JSON. No text should appear after the JSON.
    Think step by step!

    Example:
    Let's think step by step...
    |||
    {
      "target_player_name": "Charlie",
      "should_vote": true,
      "confidence": 0.75
    }
    |||

    Example to skip voting:
    Let's think step by step...
    |||
    {
      "target_player_name": null,
      "should_vote": false,
      "confidence": 0.6
    }
    |||

    Required fields:
    - target_player_name: string or null (player name, or null to skip)
    - should_vote: boolean (true to vote for target, false to skip)
    - confidence: number 0.0-1.0 (confidence in your voting decision)

  non_spy: |
    {base_context}

    You are a Villager deciding who to vote for in the Accusation Vote phase (Phase 3).
    You can skip voting if you are unsure!

    **Response Format Requirements:**
    You must respond with only valid JSON wrapped in triple pipes |||...|||
    Think through your approach before outputting JSON. No text should appear after the JSON.
    Think step by step!

    Example:
    Let's think step by step...
    |||
    {
      "target_player_name": "Charlie",
      "should_vote": true,
      "confidence": 0.88
    }
    |||

    To skip voting:
    Let's think step by step...
    |||
    {
      "target_player_name": null,
      "should_vote": false,
      "confidence": 0.45
    }
    |||

    Required fields:
    - target_player_name: string or null (player name, or null to skip)
    - should_vote: boolean (true to vote for target, false to skip)
    - confidence: number 0.0-1.0 (confidence in your voting decision)
\end{lstlisting}

\begin{figure}[htbp]
    \centering
    \includegraphics[width=\linewidth]{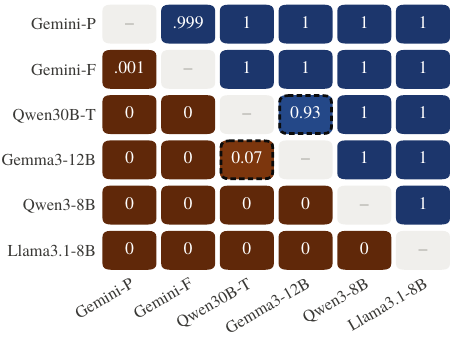}
    \caption{Probability that the row model has a higher Bradley-Terry rating than the column model, from 1000 bootstrap resamples of all matches. Values $\geq$0.975 (or $\leq$0.025) indicate a significant ordering at the 5\% level; dashed outlines mark the only pair that is not significant.}
    \label{fig:pairwise_ci}
\end{figure}

\begin{figure*}[htbp]
    \centering
    \includegraphics[width=\textwidth]{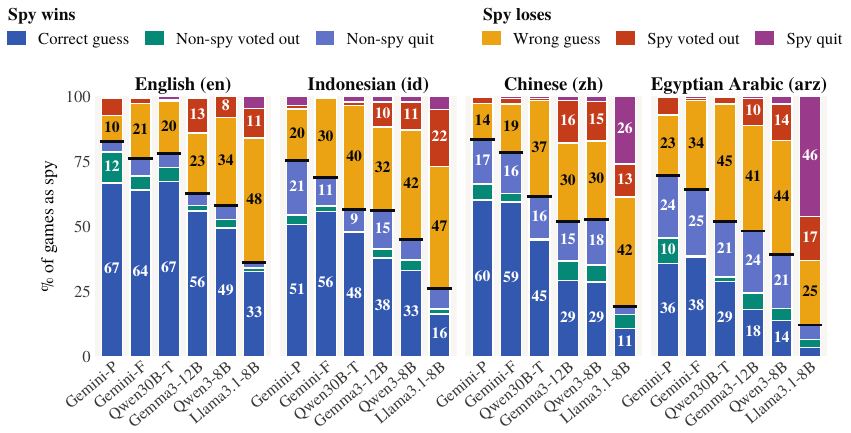}
    \caption{How games end when each model plays the spy, by language (150 games per model for \texttt{en}, 450 for each other language, pooled over the G, L and F scenarios). The black tick separates spy wins (below) from spy losses (above). No game reached the turn limit.}
    \label{fig:spy_granular_ending_language}
\end{figure*}

\section{Statistical Robustness in Overall Game Rating} \label{sec:stat_robust}

\red{In Table~\ref{tab:bt_elo_rates}, to assess rank stability under sampling variability, we compute 95\% confidence intervals with 1000-replicate match-level bootstrap following the work of \citet{chiang2024chatbotarenaopenplatform}. The top two (Gemini-P, Gemini-F) and bottom one (Llama3.1-8B) maintain non-overlapping CIs against all other models across every scenario, indicating robust positioning. However, the mid-tier models ordering between Qwen30B-T (1016) and Gemma3-12B (1003) is not statistically distinguishable in the overall ranking (CIs [1005, 1028] and [992, 1014] overlap; bootstrap $P(\text{Qwen30B-T} > \text{Gemma3-12B}) = 0.93$; Figure~\ref{fig:pairwise_ci}), and these two models flip ranks in 4 of 10 scenarios, which shows that these models are competitive. Within-scenario CIs (half-width ≈35) are approximately 3x wider than overall (≈11), reflecting the 10x smaller sample size.}

\section{\red{Why do models maintain similar rating gaps despite these spy guess accuracy drop?}} 

\begin{figure}[htbp]
    \centering
    \includegraphics[width=\linewidth]{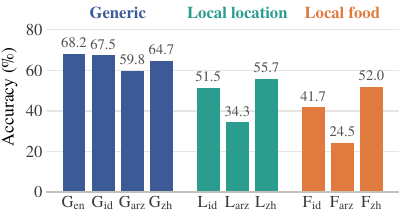}
    \caption{Spy guess accuracy (\%) by scenario, excluding games with leakage.}
    \label{fig:spy_guess_rate_by_scenario}
\end{figure}

\label{apx:why-similar}
First, we direct you to observe the following:
\begin{enumerate}
    \item In Figure~\ref{fig:spy_guess_rate_by_scenario}, in \texttt{arz}, the spy guess accuracy drops from Generic to Local Location to Local Food ($\sim$60\% $\rightarrow$ $\sim$34\% $\rightarrow$ $\sim$24\%).
    \item Focusing on \texttt{id} in Table~\ref{tab:dataset_acc_entropy} shows that entity-level uncertainty increases dramatically compared to the generic scenario: Generic entities show 67.1\% accuracy with an entropy of 1.49, while Local Food drops to 40.9\% accuracy with an entropy of 2.35. Local Locations fall to 49.6\% accuracy with an entropy of 1.81, indicating much higher uncertainty.
\end{enumerate}

These findings show that entity knowledge differentially affects model performance, but the rating shown in Table~\ref{tab:bt_elo_rates} is maintained. We dive in to answer this occurrence.

Complementing our analysis in Section~\ref{sec:cultural-knowledge} (\textbf{Lower-Accuracy Entities Share Characteristics with Distractors}), we analyzed the win rates mechanism by observing \texttt{arz} (the most challenging scenarios for a spy to guess). Below, we present the analysis of spy wins via correct guesses and non-spy wins via incorrect spy guesses.

\begin{table*}[t]
\centering
\small
\setlength{\tabcolsep}{4pt}
\begin{tabular}{lccc}
\toprule
\textbf{Model} & \makecell{\textbf{Generic}\\\textbf{Spy Win}} & \makecell{\textbf{Local Food}\\\textbf{Spy Win}} & \makecell{\textbf{Local Loc.}\\\textbf{Spy Win}} \\
\midrule
Gemini-P     & 49.33\% (n=74) & 24.67\% (n=37) & 33.33\% (n=50) \\
Gemini-F     & 47.33\% (n=71) & 28.67\% (n=43) & 38.67\% (n=58) \\
Qwen30B-T    & 50.00\% (n=75) & 15.33\% (n=23) & 21.33\% (n=32) \\
Gemma3-12B   & 33.33\% (n=50) & 10.67\% (n=16) & 10.00\% (n=15) \\
Qwen3-8B     & 22.00\% (n=33) &  8.67\% (n=13) & 10.67\% (n=16) \\
Llama3.1-8B &  7.33\% (n=11) &  1.33\% (n=2)  &  1.33\% (n=2)  \\
\bottomrule
\end{tabular}
\caption{Spy win rates with guessed location across Generic, Local Food, and Local Location settings.}
\label{tab:spy-win-apx-1}
\end{table*}

\begin{table*}[t]
\centering
\small
\setlength{\tabcolsep}{4pt}
\begin{tabular}{lccc}
\toprule
\textbf{Model} & \makecell{\textbf{Generic}\\\textbf{Non-Spy Win}} & \makecell{\textbf{Local Food}\\\textbf{Non-Spy Win}} & \makecell{\textbf{Local Loc.}\\\textbf{Non-Spy Win}} \\
\midrule
Gemini-P     & 34.00\% (n=51) & 60.00\% (n=90) & 50.67\% (n=76) \\
Gemini-F     & 23.33\% (n=35) & 54.00\% (n=81) & 44.00\% (n=66) \\
Qwen30B-T    & 28.00\% (n=42) & 54.67\% (n=82) & 50.00\% (n=75) \\
Gemma3-12B   & 30.00\% (n=45) & 43.33\% (n=65) & 35.33\% (n=53) \\
Qwen3-8B     & 20.67\% (n=31) & 59.33\% (n=89) & 38.00\% (n=57) \\
Llama3.1-8B &  4.67\% (n=7)  &  4.67\% (n=7)  &  2.67\% (n=4)  \\
\bottomrule
\end{tabular}
\caption{Non-Spy win rates (spy guessed wrong) across Generic, Local Food, and Local Location settings.}
\label{tab:nonspy-win-2}
\end{table*}

By observing Table~\ref{tab:spy-win-apx-1} and Table~\ref{tab:nonspy-win-2}, The rating stability persists because when models play as spies, their insufficient cultural knowledge hurts them, but when they play as non-spies, these occurrences help them by causing opponent spies to fail. For instance:

\begin{enumerate}
    \item In Gemini-P, spy wins drop from Generic 49.3\% $\rightarrow$ Food 29.7\%, while Non-spy wins increase (Generic 34\% $\rightarrow$ Food 60\%)
    \item In Qwen30B-T, spy wins drop from Generic 50.0\% $\rightarrow$ Food 15.3\%, while Non-spy wins increase (Generic 28\% $\rightarrow$ Food 54.7\%)
\end{enumerate}

This reveals that non-spies don't necessarily perform better strategically in local contexts, they simply benefit from opponent spy failures on unfamiliar entities. 

Another interesting thing that we observe is that spy guess accuracy does impact models’ rating placements. However, examining the above Generic scenarios in \texttt{arz} reveals that Qwen30B-T and Gemini-P achieve nearly equal spy guess rates ($\sim$50\%), yet Gemini-P maintains a higher overall rating (1136 vs 1016). This gap persists for several additional reasons, such as: Gemini-P demonstrates better spy detection capabilities when playing as a non-spy (Table~\ref{tab:nonspy_win_methods} and Figure~\ref{fig:non_spy_detective_rate}) and employs vote deception as a spy that misleads as a non-spy.
 
\section{Additional Behavior Analysis} \label{apx:add-behv}

\begin{table}[t]
\centering
\footnotesize
\setlength{\tabcolsep}{6pt}
\begin{tabular}{@{}lcccc@{}}
\toprule
\textbf{Model} & \textbf{id} & \textbf{zh} & \textbf{arz} & \textbf{Avg} \\
\midrule
Gemini-F & 2.0 & 3.3 & 0.7 & 2.0 \\
Gemini-P & 3.8 & 6.0 & 9.8 & 6.5 \\
Gemma3-12B & 3.6 & 7.6 & 6.4 & 5.9 \\
Llama3.1-8B & 2.0 & 5.3 & 3.1 & 3.5 \\
Qwen30B-T & 0.2 & 0.4 & 1.6 & 0.7 \\
Qwen3-8B & 4.0 & 6.4 & 4.7 & 5.0 \\
\bottomrule
\end{tabular}
\caption{Vote wrong rate (\%) by model across languages when these models play as a spy. This metric shows how often the opposing non-spy players incorrectly voted out a teammate. Avg is the unweighted mean across languages.}
\label{tab:vote_wrong_by_model}
\end{table}

\paragraph{Gemini-P, followed by Gemma3-12B, has the highest percentage of incorrect spy voting by non-spies} As shown in Table~\ref{tab:vote_wrong_by_model}, when Gemini-P plays the spy, non-spies incorrectly vote out a teammate most often (6.5\% on average), followed by Gemma3-12B (5.9\%), showing that these models are more subtle in concealing their role and mislead the non-spies into voting wrongly. Meanwhile, Qwen30B-T has the lowest rate (0.7\%), followed by Gemini-F (2.0\%), indicating that non-spies rarely vote out a teammate when these models play the spy.

\paragraph{The Game Dominantly Ends in one turn in Free Cycle} Overall, the game is ended on either in $P_{SG}$ or $P_{V}$ on the first turn of the Free Cycle, with mean of 6.03, standard deviation of 0.67 and median of 6.00. We attribute this due to the Round-robin Cycle which allows Spy and Non-Spy gauge the information better. However, there are a significant amount of wrong spy guess ($\approx$32.41\%) shows that the spy players are confident despite guessing wrongly and the prompt that tells them to assure to guess if they are sure.

\paragraph{Spy Guess action dominates the game ending of a spy} We then analyze the spy guess rate without leakage, where the spy must deduce the entity based on the QA history only. As shown in Figure~\ref{fig:spy_guess_rate_by_scenario}, the spy wins are mostly due to spy guess $\mathcal{E}_t$ correctly, whereas spy also loses due to spy guessing wrongly. Despite having the prompt that tell the spy to avoid guessing unless it is confident to guess, the spy tends to guess early which results also in high wrong guess (attributed more than 15\% of total games for all models except for Gemini-P in English and Chinese data). Additionally, Qwen30B-T has the highest spy guess ratio (correct or wrong) demonstrate its behavior in this game to often do this action.

\begin{figure}[htbp]
    \centering
    \includegraphics[width=\linewidth]{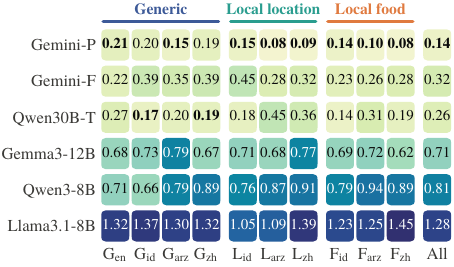}
    \caption{Non-spy vote entropy by model and scenario; lower means the non-spies agree more. \textbf{Bold}: lowest per column.}
    \label{fig:non_spy_entropy}
\end{figure}

\paragraph{Non-spy Vote Entropy} In order to measure how united the non-spies are, we check the entropy of votes across non-spy players, where these are played by the same language model can be seen in Figure~\ref{fig:non_spy_entropy}.

\paragraph{Majority wins of non-spies are due to spy guessing wrongly} Table~\ref{tab:nonspy_win_methods} shows that, overall, most non-spy wins are due to the spy guessing wrongly (74.5\% across models), while comparable percentages are due to votes caught and spies quitting. There are some differences in the Gemini family, where both have close to 60\% of non-spy wins due to the spy guessing wrongly and around 25\% due to votes caught, which is significantly higher compared to other models (around 20\%). This shows that the Gemini family have better spy detection capabilities compared to other models.

\paragraph{Spy Win Rate's rank follows overall rating} Figure~\ref{fig:win_rate_scenario} shows the model wise win rate when playing as a spy across scenarios, where it is shown that the overall spy win rate follows the trend as the overall rating in Table~\ref{tab:elo_ratings}.

\paragraph{Rank orders correspond to model size} The ranking results suggest that, based on the size and capability of the models, closed-source models outperform open-source models, and larger models outperform smaller ones (Table~\ref{tab:elo_ratings}). Additionally, within comparably sized models, we can see that Qwen3-8B outperforms Llama3.1-8B by a large margin in our benchmark, which follows a similar trend observed in other benchmarks; for instance, in the Qwen3 technical report~\cite{yang2025qwen3technicalreport}. It is worth noting that the win rate also follows the Bradley-Terry rating ranking.

\paragraph{Spy Players does not always guess correctly in match with leakage}  As shown in Table~\ref{tab:spy_guess_with_leakage}, despite the leakage happening, the spy guess rate with leakage does not guarantee to be 100\% correct, though, overall, it is still considered high (more than 80\%). Counting models that have at least 10 matches with leakage, ZH has overall lower spy guess accuracy compared to EN and ID languages which have comparable guess rate, shown by juxtaposing Gemma and Qwen models. An example of this case, given Gemini-F as Spy Model in F$_{\text{id}}$, where $\mathcal{E}_t$ is ``Bika Ambon'' (A cake from Indonesia), in one of the match, one of the non-spy leaks the entity by answering ``Can we meet Bika Ambon in festival places in Java?'' (Translated from \texttt{id}). The spy guesses wrongly by answering ``Nastar'' instead of ``Bika Ambon'' (Also a cake from Indonesia).


\begin{table}[t]
    \centering
    \small
    \setlength{\tabcolsep}{3pt}
    \caption{Spy guess accuracy with non-spy leakage by language.}
    \label{tab:spy_guess_with_leakage}
    \begin{tabular}{lcccc}
        \toprule
        Model     & en                         & id        & zh        & arz      \\
        \midrule
        Gemini-F  & 100.0                 (31) & 94.1(17)  & 92.9(14)  & 87.5(8)  \\
        Gemini-P  & 95.2                  (21) & 100.0(23) & 100.0(16) & 100.0(6) \\
        Gemma3-12B & 93.8                  (16) & 94.4(18)  & 85.7(14)  & 100.0(4) \\
        Llama3.1-8B  & 66.7                  (6)  & 100.0(5)  & 83.3(6)   & --       \\
        Qwen30B-T  & 88.5                  (26) & 95.8(24)  & 86.7(15)  & 83.3(6)  \\
        Qwen3-8B   & 88.0                  (25) & 96.3(27)  & 80.0(10)  & 0.0(1)   \\
        \bottomrule
    \end{tabular}
\end{table}

\begin{figure}[htbp]
    \centering
    \includegraphics[width=\linewidth]{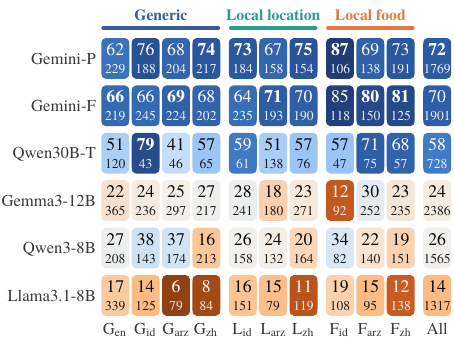}
    \caption{Non-spy voting accuracy (\%): the share of non-skip votes that target the spy, with the number of votes below. Colors are centered at the 25\% random baseline (blue: above, orange: below). \textbf{Bold}: best model per column.}
    \label{fig:non_spy_detective_rate}
\end{figure}

\begin{table}[t]
\centering
\small
\begin{tabular}{clcc}
\toprule
\textbf{Rank} & \textbf{Model} & \textbf{Rating} & \textbf{Win Rate (\%)} \\
\midrule
\textbf{1} & Gemini-P & 1135.8 & 70.13 \\
\textbf{2} & Gemini-F & 1107.4 & 66.00 \\
\textbf{3} & Qwen30B-T & 1015.7 & 51.93 \\
4 & Gemma3-12B & 1003.0 & 49.93 \\
5 & Qwen3-8B & 966.8 & 44.30 \\
6 & Llama3.1-8B & 771.3 & 17.70 \\
\bottomrule
\end{tabular}
\caption{Bradley-Terry ratings and overall win rates for each model.}
\label{tab:elo_ratings}
\end{table}

\paragraph{Non-spy Voting Accuracy} As the non-spy objective is to find the spy, we measure the performance in which the non-spy successfully identify the spy using Voting Accuracy Metric. As the number of players in the game is five, the random baseline is 25\%. Figure~\ref{fig:non_spy_detective_rate} shows the voting accuracy of each model across scenarios. Gemma3-12B, Qwen3-8B, and Llama3.1-8B have voting accuracy lower than random baseline in overall where Llama3.1-8B has the lowest voting accuracy (14\%), meanwhile, Qwen30B-T has a moderate voting accuracy and Gemini family has comparable and higher voting accuracy (around 70\%).

\paragraph{Qwen30B-T has the least voting frequencies} This is demonstrated in Figure~\ref{fig:non_spy_detective_rate}, where in the calculation SKIP vote is excluded. the skip rate of Qwen30B-T is 70.42\% where others are around (22\%-34\%). This shows that Qwen30B-T is more conservative in voting, which may lead to a higher voting accuracy.

\paragraph{Egyptian Arabic Local Entities are the hardest to be guessed by the spy} As shown in Table~\ref{tab:dataset_acc_entropy}, in both local location and food scenarios, Egyptian Arabic language has the lowest overall entity accuracy and the highest guess entropy compared to other languages, showing that these scenarios are more challenging for the spy to deduce the entity. Additionally, Indonesian local location and food scenarios have a moderate difficulty, while Chinese local location and food scenarios are the easiest for the spy to deduce the entity.

\paragraph{Spy game ending} As shown in Figure~\ref{fig:spy_granular_ending_language}, in all languages, most models tend to win through correct spy guessing. Generally, the incorrect spy guess in \texttt{id}, \texttt{zh}, and \texttt{arz}, is higher than \texttt{en}. Additionally, in \texttt{zh} and \texttt{arz} languages, all models tend to win more through non-spy voting majority compared to \texttt{en} and \texttt{id}. 

\section{Consistency Check}

\label{apx:cons-check}

\begin{table*}[t]
\centering
\small
\setlength{\tabcolsep}{5pt}
\begin{tabular}{l ccc ccc ccc}
\toprule
 & \multicolumn{3}{c}{Spy win (\%)} & \multicolumn{3}{c}{Non-spy win (\%)} & \multicolumn{3}{c}{BT rating} \\
\cmidrule(lr){2-4} \cmidrule(lr){5-7} \cmidrule(lr){8-10}
Model & R1 & R2 & R3 & R1 & R2 & R3 & R1 & R2 & R3 \\
\midrule
Gemini-P  & 60.6 & 64.8 & 52.1 & 81.8 & 85.5 & 81.8 & 1133 & 1157 & 1091 \\
Gemini-F  & 53.0 & 38.6 & 49.4 & 85.4 & 82.0 & 86.5 & 1137 & 1067 & 1121 \\
Gemma3-12B  & 39.6 & 42.5 & 37.7 & 64.6 & 64.6 & 71.7 & 1010 & 1013 & 1025 \\
Llama3.1-8B   & 11.3 &  8.5 &  9.4 &  4.9 & 14.6 & 15.4 &  638 &  701 &  715 \\
Qwen30B-T & 43.0 & 43.8 & 39.7 & 68.5 & 75.0 & 72.2 & 1029 & 1044 & 1025 \\
Qwen3-8B    & 43.2 & 38.7 & 37.8 & 75.5 & 70.9 & 72.7 & 1054 & 1018 & 1023 \\
\midrule
ICC(2,1)  & \multicolumn{3}{c}{0.93} & \multicolumn{3}{c}{0.98} & \multicolumn{3}{c}{0.97} \\
Mean SD across runs & \multicolumn{3}{c}{3.8} & \multicolumn{3}{c}{3.3} & \multicolumn{3}{c}{25} \\
\bottomrule
\end{tabular}
\caption{Consistency on F$_{\text{arz}}$: win rate as spy and as non-spy, and Bradley-Terry (BT) rating. R1, R2, R3 = three independent repeats. Results obtained on the 598 games. ICC(2,1) = intraclass correlation (two-way random, absolute agreement) across models; mean SD = average per-model standard deviation across runs (percentage points for win rates, rating points for ratings).}
\label{tab:consistency-run}
\end{table*}

To check the consistency of the evaluation framework, we perform three repeated runs on the Egyptian Food entity, as it is the most difficult one and may show the biggest inconsistency. Due to budget constraints, we run 598 games two more times, with the same matches tested across runs.

Table~\ref{tab:consistency-run} shows that our results are reliable. Measured by the Intraclass Correlation Coefficient (ICC), the spy and non-spy win rates are 0.93 and 0.98. The Bradley-Terry rating is also stable at 0.97. The difference between models is much larger than the difference between runs (SD 161 vs.\ 25 rating points).

In every run, the models always fall into the same performance levels (Gemini $>$ Qwen or Gemma $>$ Llama). Rankings only change within these levels, where the confidence intervals overlap.

While the overall scores are stable, individual games are much more stochastic. Across all three runs, we saw the exact same winner in only 62\% of the games (Fleiss' $\kappa = 0.47$; pairwise agreement 73--76\%).

\section{Guidelines for Selecting Local Locations and Foods}
\label{appendix:entity-selection}

\paragraph{Goal and Scope}
The goal is to construct separate pools of approximately 30 recognizable local locations and 30 foods for a target country, region, or community. In our experiments, each pool contains 30 entities, as listed in Appendix~\ref{appendix:entities}. These guidelines concern the Local Location (L) and Local Food (F) scenarios; the Generic (G) scenario uses the original game's generic locations and their translations.

All players see the same candidate list, while only non-spies know which entity has been selected for a game. Candidates should therefore be familiar by name and evoke a shared understanding that players can communicate indirectly. A location is a real, named place that members of the target community can imagine visiting. This includes recognizable establishments and chains, as well as individual landmarks, institutions, and areas. A food is a named dish, snack, dessert, or beverage associated with the community's everyday life or traditions.

\subsection{Step 1: Brainstorm by Category}

Start with categories that capture varied aspects of the target culture, then brainstorm several candidates within each category. Categories are suggestions rather than fixed quotas and may be adapted to the cultural setting.

Locations may cover restaurants and caf\'es, shopping areas, landmarks, leisure destinations, educational institutions, religious sites, and neighborhoods. Foods may cover everyday meals, street snacks, breads, desserts, beverages, and celebratory dishes.

\subsection{Step 2: Check Each Candidate}

\paragraph{Mental Picture Test}
Does the name evoke a recognizable experience? For a location, consider its indoor or outdoor setting, typical activities, crowding, atmosphere, cost, and visitors. For a food, consider its ingredients, taste, texture, preparation, serving temperature, presentation, and the settings or occasions in which it is consumed. A useful candidate supports several shared associations beyond recognition of its name alone.

\paragraph{Question Test}
Can players discuss the entity without naming it directly? Check whether at least three to five attributes could support indirect questions. For locations, these might concern noise, visiting hours, typical visitors, dress, purchases, or activities. For foods, they might concern whether the item is eaten hot or cold, sweet or savory, as a meal or a snack, with utensils or by hand, or on everyday versus special occasions. Writing a fixed question set is unnecessary; the purpose is to ensure that meaningful indirect discussion is possible.

\paragraph{Confusability Test}
Does the candidate share features with at least two or three other entities in the same pool? Similar but distinct candidates create clusters that require reasoning over subtle differences. For example, ITB, Binus, and SMAN 8 Jakarta form an educational cluster in the Indonesian location pool; \zh{北京大学} (Peking University), \zh{清华大学} (Tsinghua University), and \zh{复旦大学} (Fudan University) provide a Chinese example. The Egyptian pool similarly includes the American University in Cairo, Nile University, and Ain Shams University. Food clusters include the Indonesian beverages Bandrek, Beras Kencur, and Loloh Cemcem; the Chinese breads \zh{馒头} (Mantou), \zh{花卷} (Huajuan), and \zh{窝头} (Wotou); and the Egyptian beverages \arabictext{ينسون} (Yansoon), \arabictext{كركديه} (Karkadeh), and \arabictext{سحلب} (Sahlab). These examples illustrate shared features while retaining differences that can support questioning. Consider replacing isolated candidates that a single broad hint would identify immediately.

\paragraph{Familiarity Test}
Would most members of the intended community recognize the candidate by name, even without having visited the place or consumed the food? Prefer widely known entities over obscure establishments, streets known only within a small neighborhood, or dishes and variants familiar only to a narrow group. Familiarity should be assessed relative to the stated geographic and cultural scope rather than international visibility.

\subsection{Step 3: Build a Balanced Final Pool}

After filtering, choose approximately 30 entities for each pool. Balance category coverage with confusability: each cluster should contain at least two or three entities, while the overall list should span varied settings, activities, or culinary experiences. Recheck the final pool because removing a candidate can leave another without plausible alternatives. Use recognizable local names and avoid duplicate entries that are merely alternative names for the same entity. The aim is a varied sample of local culture that supports both shared understanding and meaningful uncertainty during gameplay.


\end{document}